\documentclass[11pt]{article}

\usepackage[preprint]{acl}

\usepackage{times}
\usepackage{latexsym}
\usepackage{multirow}
\usepackage[T1]{fontenc}

\usepackage[utf8]{inputenc}

\usepackage{microtype}

\usepackage{inconsolata}
\usepackage{url}
\usepackage{graphicx}

\usepackage{amsmath}
\usepackage{amssymb}
\usepackage{booktabs}
\title{De-conflating Preference and Qualification: Constrained Dual-Perspective Reasoning for Job Recommendation with Large Language Models}

\author{
  \textbf{Bryce Kan$^{1*}$ \quad
  Wei Yang$^{1*}$ \quad
  Emily Nguyen$^{1*}$ \quad
  Ganghui Yi$^{1}$} \\
  \textbf{Bowen Yi$^{1}$ \quad
  Chenxiao Yu$^{1}$ \quad
  Yan Liu$^{1}$} \\
  $^{1}$University of Southern California \\
  \{brycekan, wyang930, emilyn98, yanliu.cs\}@usc.edu
}

\begin{document}
\maketitle
\begingroup
\renewcommand\thefootnote{\fnsymbol{footnote}}
\footnotetext[1]{Equal contribution.}
\endgroup

\begin{abstract}
Professional job recommendation involves a complex bipartite matching process that must reconcile a candidate’s subjective preference with an employer’s objective qualification. While Large Language Models (LLMs) are well-suited for modeling the rich semantics of resumes and job descriptions, existing paradigms often collapse these two decision dimensions into a single interaction signal, yielding confounded supervision under recruitment-funnel censoring and limiting policy controllability. To address these challenges, We propose \textbf{JobRec}, a generative job recommendation framework for \textbf{de-conflating preference and qualification} via constrained dual-perspective reasoning. JobRec introduces a Unified Semantic Alignment Schema that aligns candidate and job attributes into structured semantic layers, and a Two-Stage Cooperative Training Strategy that learns decoupled experts to separately infer preference and qualification. Building on these experts, a Lagrangian-based Policy Alignment module optimizes recommendations under explicit eligibility requirements, enabling controllable trade-offs. To mitigate data scarcity, we construct a synthetic dataset refined by experts. Experiments show that JobRec consistently outperforms strong baselines and provides improved controllability for strategy-aware professional matching. The code is available at \url{https://github.com/brycekan123/DualOptimization_jobrec}.

\end{abstract}


\section{Introduction}

Online recruitment platforms rely on recommender systems to personalize bipartite matching between candidates and employers~\cite{alsaif2022learning,ccelik2025job}. Traditional keyword-based methods often fail to capture the complexity of professional profiles and job requirements~\cite{mulay2022job,tavakoli2022ai,salinas2023unequal}, whereas Large Language Models (LLMs) ~\cite{chen2025tourrank,chen2026self} provide a promising foundation by modeling long-context text and fine-grained semantic relations beyond lexical overlap~\cite{wang2024mllm4rec, wang2023recmind, zhang2023recommendation}. However, resumes and job descriptions (JDs) are semantically and structurally heterogeneous, combining structured attributes such as education and experience with unstructured narratives that convey career trajectories, skill dependencies, and project evidence~\cite{du2024enhancing,sun2025market}. As a result, translating LLM semantic understanding into reliable policies that reflect labor-market decision logic remains an open challenge~\cite{hu2025fairwork,zheng2023generative}.

\begin{figure}[t]
    \centering
    \includegraphics[width=0.5\textwidth]{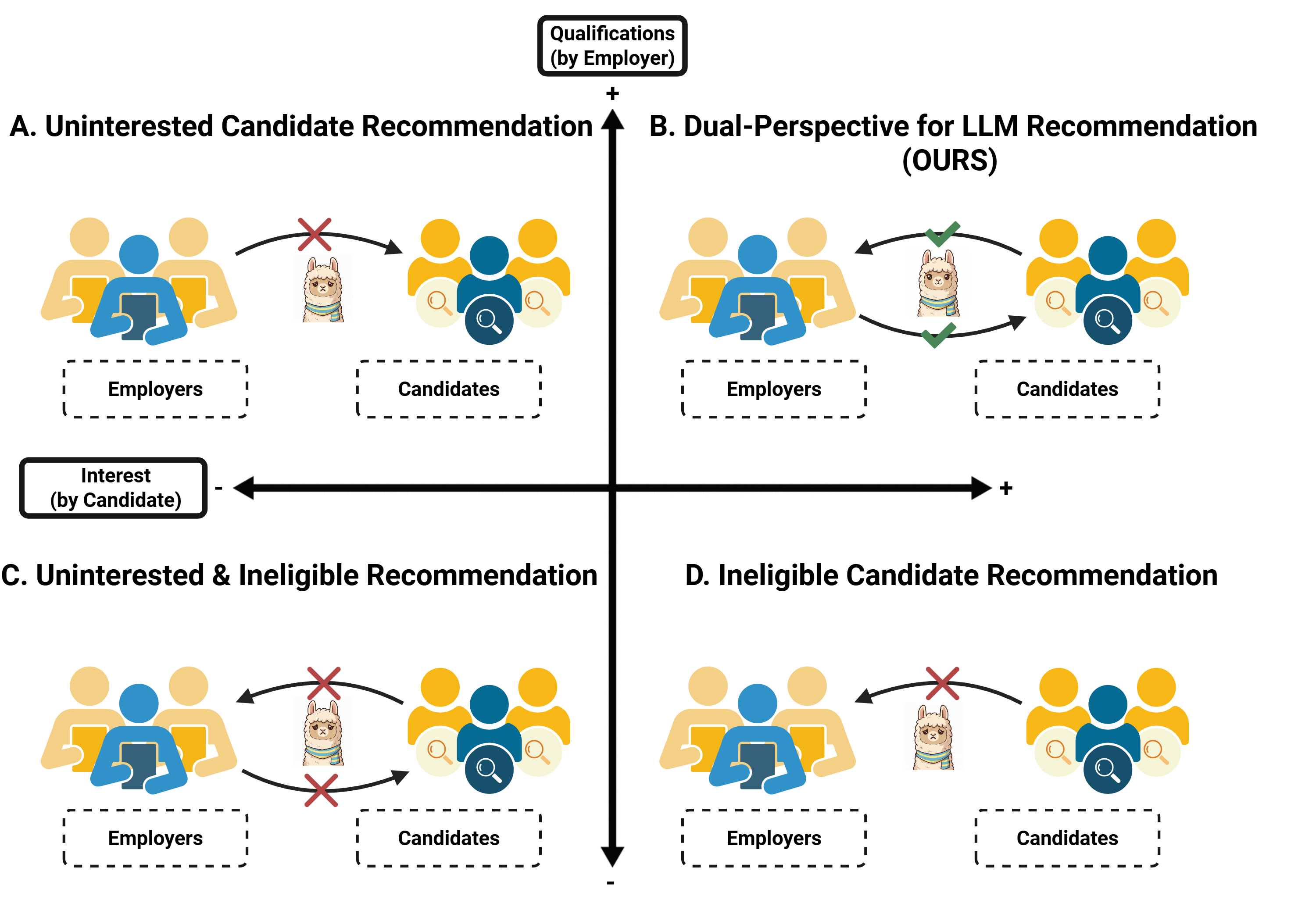}
    \caption{\textbf{Job recommendation is inherently dual-perspective}: candidates focus on preference (interest or willingness to apply) while employers enforce qualification (eligibility to be shortlisted). Effective matching should prioritize (B) "high-interest-high-eligibility" region and particularly avoid both (A) “high interest but low eligibility” and (D) “high eligibility but low interest” recommendations.}
    \label{fig:intro_architecture}
\end{figure}

Despite their strong semantic capacity, existing LLM-based methods face three fundamental challenges to achieving principled, decision-faithful matching. First, \textbf{dual-perspective signal conflation} remains pervasive in professional matching. As shown in Figure~\ref{fig:intro_architecture}, a successful outcome is jointly determined by a candidate’s subjective preference and an employer’s objective qualification~\cite{yang2022modeling}. Preference captures whether the candidate would apply, and qualification reflects whether the candidate passes employer screening. However, previous methods often collapse these heterogeneous signals into a single interaction label, such as match or hire, resulting in models that entangle aspiration with eligibility~\cite{wu2024exploring,kurek2024zero}. This simplification can lead to biased recommendations, including aspirational but infeasible roles and feasible but uninspiring options. Furthermore, this challenge is exacerbated by \textbf{counterfactual ambiguity under censored feedback}. Negative outcomes are non-diagnostic and may arise from missing exposure, disinterest, screening rejection, or dropout. Such funnel-driven censoring makes supervision noisy and confounded, hindering attribution of failure modes and robust learning of preference versus qualification boundaries.~\cite{tavakoli2022ai,salinas2023unequal}.

Additionally, job recommendation requires \textbf{policy controllability under operational eligibility considerations} rather than a static ranking score. In practice, platforms should promote vacancies aligned with candidate interests while managing the eligibility risks of low-qualification recommendations, including wasted exposure, inefficient applications, and reduced employer-side satisfaction~\cite{kokkodis2023learning,kokkodis2023good}. However, prevailing approaches typically rely on heuristic scalarization, such as fixed linear combinations, which lacks explicit eligibility semantics and offers no principled mechanism to control the trade-off between candidate-centric and employer-centric strategies. Consequently, existing systems lack a strategy-aware decision layer that adapts recommendation behavior to different operating conditions without ad hoc retuning or retraining.


To bridge these gaps, we propose \textbf{JobRec}, which \textbf{de-conflates preference and qualification} through \textbf{constrained dual-perspective reasoning}. JobRec explicitly separates the two decision processes and aligns them with a controllable policy layer. We introduce a \textbf{Unified Semantic Alignment Schema (USAS)}, a hierarchical taxonomy that symmetrically organizes candidate and job attributes into four semantic layers. Building on USAS, we develop a \textbf{two-stage cooperative training} strategy. Stage~I trains decoupled experts in an LLM backbone to infer candidate preference and employer qualification, and Stage~II applies \textbf{Lagrangian-based policy alignment} to optimize ranking under explicit eligibility requirements, enabling controllable trade-offs across operating regimes. To mitigate data scarcity, we also construct a Job synthetic dataset refined by experts to provide reliable dual-perspective supervision.

Our contributions are summarized as follows:
\begin{itemize}
    \item We propose \textbf{JobRec}, a dual-perspective LLM-based job recommendation framework that uses \textbf{USAS} to align candidate and job attributes into four semantic layers for preference--qualification de-conflation.
    \item We introduce a \textbf{Lagrangian-based constrained policy optimization} framework, enabling controllable trade-offs between preference and eligibility across operating regimes.
    \item We construct and release a CS-domain expert-refined dataset with dual-perspective annotations, providing reliable supervision and a benchmark for professional matching.
\end{itemize}

\section{Related Work}

\subsection{Generative Recommendation}
\label{sec:related_generative_rec}

Generative recommendation casts recommendation as language modeling by serializing user histories and items and solving them with pre-trained LLMs, aiming to improve transferability and cold-start robustness~\cite{li2023text,cui2022m6}. A central challenge is misalignment between LLM priors and recommender objectives, especially collaborative signals and the discrete item space. Recent work addresses this via recommendation-oriented tuning and alignment~\cite{bao2023tallrec,wang2024pre} and stronger item grounding~\cite{bao2025bi,tan2024idgenrec}. Practical systems also target efficient inference through ranking-centric designs that reduce or avoid long-form generation~\cite{wang2024rethinking,yue2023llamarec}. Beyond end-to-end generation, LLMs increasingly act as augmenters or interfaces that enhance conventional recommenders through representation alignment, collaborative knowledge injection, and tool-based decision pipelines~\cite{wei2024llmrec,ren2024representation,kim2024large,zhang2024notellm,zhang2024notellm2}.

\subsection{Job Recommendation}
Job recommender systems match candidates with vacancies under heterogeneous and incomplete signals, with stricter eligibility constraints and higher-stakes outcomes than general recommendation~\cite{ccelik2025job}. 
Early work relies on content-based matching from explicit profile attributes~\cite{alsaif2022learning} and hybrid models that combine content with collaborative signals to mitigate sparsity~\cite{mulay2022job}. 
Recent studies incorporate market signals and long-horizon utility, such as demand-aware skill pathway recommendation~\cite{tavakoli2022ai} and explainable multi-objective reinforcement learning~\cite{sun2025market}. 
LLMs further enable richer semantic profiling and reasoning, including alignment-based resume refinement~\cite{du2024enhancing,wu2024exploring}, and zero-shot job--candidate matching with pretrained encoders~\cite{kurek2024zero}. 
Fairness is central in hiring, and documented intersectional demographic biases motivate systematic auditing and mitigation~\cite{salinas2023unequal}. The full related work is provided in Appendix~\ref{app:appendix_relatedwork}.

\section{Methodology}
\label{sec:methodology}
\begin{figure*}[t]
    \centering
    \includegraphics[width=\textwidth]{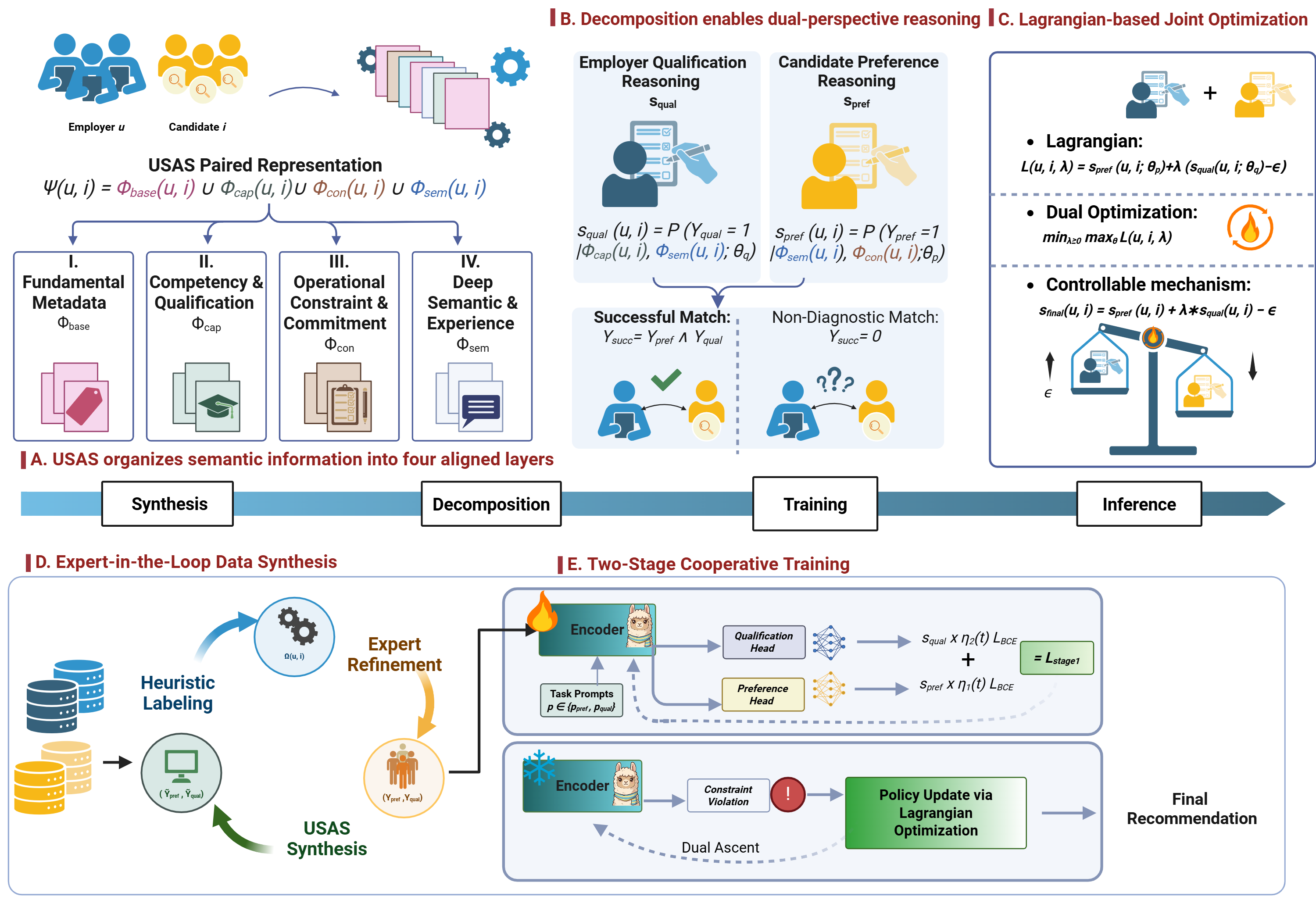}
    \caption{The architecture of our framework, consisting of (A) \textbf{Unified Semantic Alignment Schema (USAS)} for symmetric candidate--job feature alignment, (B) \textbf{Problem Decomposition: Preference versus Qualification} to obtain disentangled scores $s_{\text{pref}}$ and $s_{\text{qual}}$, (C) \textbf{Lagrangian-based Joint Optimization} to derive a constraint-aligned ranking policy, (D) \textbf{Expert-in-the-Loop Data Synthesis} to construct dual-perspective supervision, and (E) \textbf{Two-Stage Cooperative Training} to learn disentangled experts and perform policy alignment.}
    \label{fig:overall_architecture}
\end{figure*}

In this section, we will introduce the main components of our method. The overall architecture is shown in Figure~\ref{fig:overall_architecture}.

\subsection{Unified Semantic Alignment Schema}
\label{sec:usas}


Job recommendation requires a representation that can simultaneously encode (i) heterogeneous structured attributes (e.g., education, location, availability) and (ii) unstructured narratives (e.g., project evidence and role descriptions) in a symmetrically aligned manner for both candidates and jobs. To this end, we propose the \textbf{Unified Semantic Alignment Schema (USAS)}, a hierarchical schema that organizes recruitment information into four aligned layers. For a candidate $u$ and a job $i$, USAS yields a paired representation
\begin{equation}
\begin{aligned}
\Psi(u,i)
&= \Phi_{base}(u,i) \cup \Phi_{cap}(u,i) \\
&\quad \cup \Phi_{con}(u,i) \cup \Phi_{sem}(u,i)
\end{aligned}
\end{equation}
where each layer is defined on both sides (candidate or job) to enable consistent comparison and subsequent dual-perspective reasoning.

\paragraph{(I) Fundamental Metadata Layer ($\Phi_{base}$).}
This layer encodes coarse identifiers and categorical anchors. For candidates, $\mathcal{U}_{base}=\{id_u,\text{row}_u,\text{lvl}_u\}$ (e.g., academic level $\text{lvl}_u$); for jobs, $\mathcal{I}_{base}=\{id_i,\text{ind}_i,\text{min\_lvl}_i\}$. These anchors provide global context and support coarse-grained candidate--vacancy compatibility.

\paragraph{(II) Competency \& Qualification Layer ($\Phi_{cap}$).}
This layer captures eligibility-related credentials and screening criteria. Candidate qualifications are represented as $\mathcal{U}_{cap}$, including academic metrics (e.g., $\text{GPA}\in[0,4.0]$, $\text{GPA}_{ug}$), disciplinary specializations (e.g., $\text{major}_{master},\text{major}_{second}$), and research output (e.g., publications $N_{pub}$). Job-side eligibility is represented as $\mathcal{I}_{cap}$ with threshold-like criteria $\mathcal{B}=\{\tau_{gpa},\tau_{exp},\mathcal{M}_{acc}\}$, where $\mathcal{M}_{acc}$ denotes acceptable majors. 

\paragraph{(III) Operational Constraint \& Commitment Layer ($\Phi_{con}$).}
This layer models practical compatibility constraints that often govern whether a match is actionable. For candidates, we encode availability and commitments (e.g., weekly bandwidth $h_u$, seasonal availability $c_{summer}\in\{0,1\}$, and location modality $m_{loc}$). For jobs, we encode operational demands (e.g., required bandwidth $h_i$ and modality $m_{job}$), ensuring that preference reasoning is grounded in operational viability.

\paragraph{(IV) Deep Semantic \& Experience Layer ($\Phi_{sem}$).}
To capture fine-grained evidence beyond structured fields, we extract semantically rich representations from narratives. On the candidate side, $\mathcal{U}_{sem}$ includes project and experience descriptors $P=\{proj_k\}_{k=1}^{3}$ and $E=\{ext_k\}_{k=1}^{2}$, augmented by an LLM-synthesized persona embedding $\mathbf{z}_u$. On the job side, $\mathcal{I}_{sem}$ includes required skills $\mathcal{S}_{req}$ and an LLM-generated distillation of JD nuances $\mathbf{z}_i$.


\subsection{Problem Decomposition: Preference versus Qualification}
\label{sec:decomposition}
In bipartite professional matching, success depends on two distinct decisions: whether a candidate is interested in a job and whether the candidate meets employer-side screening requirements. Most existing approaches collapse these dimensions into a single interaction signal such as hire or match, which entangles aspiration with eligibility. This is especially problematic under recruitment-funnel censoring, where unobserved intermediate stages such as exposure, application, and screening make negative outcomes non-diagnostic. Building on USAS, we decompose job recommendation into two disentangled reasoning tasks.

\paragraph{(I) Candidate Preference Reasoning ($s_{pref}$).}
We define $s_{pref}(u,i)\in[0,1]$ as the probability that candidate $u$ would choose to apply to job $i$:
\begin{equation}
\begin{aligned}
s_{pref}(u,i)
&= P\!\left( Y_{pref}=1 \mid \right. \\
&\qquad \left. \Phi_{sem}(u,i),\, \Phi_{con}(u,i);\, \theta_p \right)
\end{aligned}
\end{equation}
where $\theta_p$ denotes the preference module parameters. This score is primarily driven by semantic alignment (career interests and experience signals in $\Phi_{sem}$) together with practical compatibility in $\Phi_{con}$.

\paragraph{(II) Employer Qualification Reasoning ($s_{qual}$).}
We define $s_{qual}(u,i)\in[0,1]$ as the probability that candidate $u$ satisfies the screening criteria of job $i$ (conditioned on an application):
\begin{equation}
\begin{aligned}
s_{qual}(u,i)
&= P\!\left( Y_{qual}=1 \mid \right. \\
&\qquad \left. \Phi_{cap}(u,i),\, \Phi_{sem}(u,i);\, \theta_q \right)
\end{aligned}
\end{equation}
where $\theta_q$ denotes the qualification module parameters. This score is mainly determined by eligibility-related credentials in $\Phi_{cap}$ (e.g., thresholds such as $\text{GPA}\ge \tau_{gpa}$) with complementary evidence from $\Phi_{sem}$.

\paragraph{Why De-conflation Matters.}
Let $Y_{succ}$ denote the observed terminal success signal. Conceptually, a successful match requires both interest and qualification:
\begin{equation}
    Y_{succ} = Y_{pref} \wedge Y_{qual}.
\end{equation}
However, real-world logs rarely reveal $Y_{pref}$ and $Y_{qual}$ separately. When $Y_{succ}=0$, the outcome is non-diagnostic: it may arise because the candidate was never exposed to the job, chose not to apply, failed screening, or dropped out for reasons unrelated to preference or qualification. This funnel-induced censoring yields selection bias and ambiguous supervision if one trains directly on $Y_{succ}$, since the model cannot attribute a negative outcome to the correct underlying cause. By explicitly modeling $s_{pref}$ and $s_{qual}$, JobRec enables \textbf{dual-perspective reasoning} and provides the necessary ingredients for the constrained optimization in Sec.~\ref{sec:lagrangian}.


\subsection{Lagrangian-based Joint Optimization}
\label{sec:lagrangian}

After disentangling preference and qualification, we cast recommendation as a \emph{constrained optimization} problem: for each candidate $u$, we seek jobs that maximize preference while meeting a minimum eligibility level measured by qualification,
\begin{equation}
\begin{aligned}
    \max_{\theta} \quad & s_{pref}(u,i;\theta_p) \\
    \text{s.t.} \quad & s_{qual}(u,i;\theta_q) \ge \epsilon,
\end{aligned}
\end{equation}
where $\epsilon\in[0,1]$ is a screening threshold that controls the strictness of eligibility requirements. This formulation discourages recommending high-interest but low-eligibility jobs, as well as high-eligibility but low-interest ones.

\paragraph{(I) Lagrangian Transformation.}
To optimize the constrained objective in a differentiable framework, we introduce a nonnegative Lagrange multiplier $\lambda\ge 0$ and form the Lagrangian:
\begin{equation}
    \mathcal{L}(u,i,\lambda)
    = s_{pref}(u,i;\theta_p) + \lambda\big(s_{qual}(u,i;\theta_q)-\epsilon\big)
\end{equation}
Intuitively, $\lambda$ adapts the penalty for eligibility violations: when a candidate--job pair falls below the threshold, the optimization increases the emphasis on qualification to steer the policy back toward the feasible region.

\paragraph{(II) Dual Optimization and Policy Alignment.}
We optimize JobRec via the saddle-point objective
\begin{equation}
    \min_{\lambda \ge 0} \max_{\theta} \ \mathcal{L}(u,i,\lambda)
\end{equation}
which aligns the learned policy with explicit eligibility requirements. At inference time, we rank jobs using the Lagrangian-aligned score
\begin{equation}
    s_{final}(u,i) = s_{pref}(u,i) + \lambda^{*}\big(s_{qual}(u,i)-\epsilon\big)
\end{equation}
where $\lambda^{*}$ is the converged (or selected) multiplier. This design provides a controllable mechanism for switching operating regimes: increasing $\epsilon$ or $\lambda$ yields a more eligibility-centric policy, while decreasing them yields a more candidate-centric policy.

\subsection{Expert-in-the-Loop Data Synthesis}
\label{sec:data_synthesis}

High-quality supervision for professional matching is scarce, particularly labels that separate candidate preference from employer qualification. We therefore adopt an Expert-in-the-Loop protocol to synthesize coherent candidate--job pairs and annotate $(Y_{pref}, Y_{qual})$.

\paragraph{USAS-guided pair generation.}
We use USAS (Sec.~\ref{sec:usas}) as a template to synthesize candidate profiles and job descriptions with controlled diversity and internal consistency. We first estimate realistic distributions over structured fields such as education, GPA, skills, availability, and location, then sample and compose attributes across the four USAS layers to instantiate candidates and jobs. This ensures that narratives align with structured credentials and that job requirements match the role and domain context.

\paragraph{Heuristic labeling with expert refinement.}
We first assign silver labels using a rule-based scoring function $\Omega(u,i)$ that measures alignment with screening criteria and required skills. Domain experts then review and revise these labels, correcting cases that require technical judgment and calibrating borderline examples. The resulting annotations $(Y_{pref}, Y_{qual})$ provide training supervision for our disentangled experts (Sec.~\ref{sec:training}). Details of the synthetic data are provided in Appendix~\ref{app:appendix_exp_syndata}.

\subsection{Two-Stage Cooperative Training}
\label{sec:training}

To instantiate JobRec in a stable and controllable manner, we adopt a Two-Stage Cooperative Training strategy. The key idea is to learn disentangled predictors for preference and qualification, then align the final policy with explicit eligibility requirements via Lagrangian-based optimization.

\paragraph{Phase I: Multi-Task Disentangled Pre-training.}
In Stage~I, we train a shared transformer encoder with two task-specific heads to predict candidate preference and employer qualification. Given USAS-aligned inputs $\Psi(u,i)$ and task prompts $\mathbf{p}\in\{p_{pref},p_{qual}\}$, the model outputs $(s_{pref}, s_{qual})$ through two independent linear heads. To balance dense preference signals and sparser qualification labels, we use adaptive task weighting:
\begin{equation}
    \mathcal{L}_{stage1}
    = \eta_1(t)\,\mathcal{L}_{BCE}(s_{pref}) + \eta_2(t)\,\mathcal{L}_{BCE}(s_{qual})
\end{equation}
where $\eta_1(t)$ and $\eta_2(t)$ are learnable weights updated during training. This stage encourages specialization while reducing gradient interference between the two objectives.

\paragraph{Phase II: Lagrangian Policy Alignment.}
After Stage~I stabilizes, Stage~II calibrates the policy under the constrained objective in Sec.~\ref{sec:lagrangian}. We freeze the shared encoder and update only the policy layer and the dual variable. For each minibatch, we update the policy via
\begin{equation}
    \theta_{policy} \leftarrow \theta_{policy} + \alpha \nabla_{\theta}
    \left[ s_{pref} + \lambda\cdot(s_{qual}-\epsilon) \right]
\end{equation}
and update $\lambda$ with a projected dual ascent step as described in the Appendix. This stage aligns ranking with eligibility requirements and enables controllable trade-offs across operating regimes.

\begin{table*}[t]
\centering
\small
\setlength{\tabcolsep}{6pt}
\begin{tabular}{l c c c c c c c}
\toprule
\textbf{Model} & \textbf{Task} & \textbf{Recall@1} & \textbf{Recall@3} & \textbf{Recall@5} & \textbf{NDCG@1} & \textbf{NDCG@3} & \textbf{NDCG@5} \\
\midrule
\multirow{2}{*}{LLM Zero-shot} 
& Preference & 0.0849 & 0.2033 & 0.2896 & 0.0849 & 0.1532 & 0.1884 \\
& Qualification & 0.0556 & 0.1778 & 0.2667 & 0.0556 & 0.1254 & 0.1622 \\
\midrule
\multirow{2}{*}{In-Context Learning} 
& Preference & 0.2382 & 0.3081 & 0.3352 & 0.2382 & 0.2782 & 0.2898 \\
& Qualification & 0.1889 & 0.3444 & 0.3778 & 0.1889 & 0.2783 & 0.2922 \\
\midrule
\multirow{2}{*}{TallRec} 
& Preference & 0.2240 & 0.2696 & 0.3224 & 0.2240 & 0.2506 & 0.2721 \\
& Qualification & 0.1000 & 0.1667 & 0.2333 & 0.1000 & 0.1333 & 0.1591 \\
\midrule
\multirow{2}{*}{\textbf{JobRec (ours)}} 
& Preference & \textbf{0.4680} & \textbf{0.6720} & \textbf{0.7520} & \textbf{0.4680} & \textbf{0.5870} & \textbf{0.6200} \\
& Qualification & \textbf{0.3330} & \textbf{0.5330} & \textbf{0.7670} & \textbf{0.3330} & \textbf{0.4510} & \textbf{0.5480} \\
\bottomrule
\end{tabular}
\vspace{-2mm}
\caption{Performance comparison with LLM-based baselines for dual-perspective job recommendation, including both \emph{Preference} and \emph{Qualification}.
JobRec consistently outperforms all baselines across metrics, with the most pronounced improvements on qualification-aware ranking.}
\label{tab:rq1_llm_baselines_full}
\vspace{-3mm}
\end{table*}

\section{Experiments}

\subsection{Experimental Setup}
\label{sec:exp_setup}

We evaluate on a CS-domain job recommendation benchmark with $5{,}000$ candidates and $100$ job postings, where each candidate--job pair has dual annotations for Preference and Qualification. We follow a standard sampled top-$K$ retrieval protocol and report Recall@$K$ and NDCG@$K$ on both tasks. Unless otherwise specified, JobRec uses a Llama-3-8B backbone fine-tuned with LoRA, and we select checkpoints based on validation performance. We compare against LLM baselines including Zero-Shot Prompting, In-Context Learning with four demonstrations, and TallRec, as well as non-LLM recommenders including SimpleX, LightGCN, xDeepFM, DeepFM, and DCNv2. Full details are provided in the Appendix~\ref{app:appendix_exp_setup}.

\subsection{RQ1: How does JobRec compare with LLM-based recommendation baselines?}
Table~\ref{tab:rq1_llm_baselines_full} compares JobRec with prompting-based and generative LLM baselines under the same dual-perspective evaluation. Prompting alone is insufficient for recruitment-style decisions. Zero-shot prompting performs poorly on both preference and qualification, and in-context learning with four demonstrations improves results but still fails to jointly model candidate interest and employer eligibility because the supervision signal remains implicit and entangled. In contrast, JobRec achieves large and consistent gains on both perspectives. Compared to the strongest prompting baseline, LLM+ICL, JobRec increases Recall@5 from $0.3352$ to $0.7520$ on preference and from $0.3778$ to $0.7670$ on qualification, with NDCG@5 improving from $0.2898$ to $0.6200$ and from $0.2922$ to $0.5480$, respectively. The strongest gains appear on qualification across $K\in\{1,3,5\}$, consistent with our view that eligibility behaves as a structured decision boundary rather than semantic similarity. This pattern also explains TallRec, which remains competitive on preference but lags on qualification, suggesting that generative pipelines capture plausibility while failing to enforce eligibility evidence.

\begin{table*}[t]
\centering
\small
\setlength{\tabcolsep}{6pt}
\begin{tabular}{l c c c c c c c}
\toprule
\textbf{Model} & \textbf{Task} & \textbf{Recall@1} & \textbf{Recall@3} & \textbf{Recall@5} & \textbf{NDCG@1} & \textbf{NDCG@3} & \textbf{NDCG@5} \\
\midrule
\multirow{2}{*}{SimpleX}
& Preference     & 0.0164 & 0.0549 & 0.0920 & 0.0164 & 0.0383 & 0.0534 \\
& Qualification  & 0.0333 & 0.0667 & 0.1333 & 0.0333 & 0.0544 & 0.0802 \\
\midrule
\multirow{2}{*}{LightGCN}
& Preference     & 0.0228 & 0.0735 & 0.1191 & 0.0228 & 0.0518 & 0.0704 \\
& Qualification  & 0.1000 & 0.1667 & 0.1667 & 0.1000 & 0.1421 & 0.1421 \\
\midrule
\multirow{2}{*}{DeepFM}
& Preference     & 0.0257 & 0.0849 & 0.1355 & 0.0257 & 0.0594 & 0.0800 \\
& Qualification  & 0.0667 & 0.1333 & 0.1667 & 0.0667 & 0.1044 & 0.1173 \\
\midrule
\multirow{2}{*}{xDeepFM}
& Preference     & 0.0528 & 0.1056 & 0.1526 & 0.0528 & 0.0828 & 0.1022 \\
& Qualification  & 0.1000 & 0.2000 & 0.2667 & 0.1000 & 0.1631 & 0.1918 \\
\midrule
\multirow{2}{*}{DCNv2}
& Preference     & 0.0392 & 0.1006 & 0.1362 & 0.0392 & 0.0752 & 0.0898 \\
& Qualification  & 0.0667 & 0.1000 & 0.1667 & 0.0667 & 0.0833 & 0.1106 \\
\midrule
\multirow{2}{*}{\textbf{JobRec (ours)}}
& Preference     & \textbf{0.4680} & \textbf{0.6720} & \textbf{0.7520} & \textbf{0.4680} & \textbf{0.5870} & \textbf{0.6200} \\
& Qualification  & \textbf{0.3330} & \textbf{0.5330} & \textbf{0.7670} & \textbf{0.3330} & \textbf{0.4510} & \textbf{0.5480} \\
\bottomrule
\end{tabular}
\vspace{-2mm}
\caption{Performance comparison with representative deep recommendation baselines for dual-perspective job recommendation.
JobRec consistently achieves the strongest results on both preference and qualification, substantially widening the gap as $K$ increases.}
\label{tab:rq2_deep_baselines}
\vspace{-3mm}
\end{table*}

\subsection{RQ2: How does JobRec compare with general deep recommendation models?}
Table~\ref{tab:rq2_deep_baselines} compares JobRec with representative collaborative filtering and general deep recommenders. These baselines perform poorly on both tasks, most notably on preference. This limitation is expected in professional matching, where preference relies on high-level semantic alignment such as research interests, project narratives, and role intent, and qualification follows compositional, rule-like constraints such as degree, major, GPA, and experience. Both signals are difficult to learn from sparse interactions with ID-centric objectives, and the issue is amplified when supervision conflates the two perspectives. JobRec delivers large gains on both preference and qualification by combining semantic modeling with constraint-aware decision making. Against the strongest deep baseline on preference, xDeepFM, JobRec improves Recall@5 from $0.1526$ to $0.7520$ and NDCG@5 from $0.1022$ to $0.6200$. On qualification, xDeepFM reaches Recall@5 of $0.2667$, while JobRec achieves $0.7670$ with NDCG@5 $0.5480$. The gap increases with larger $K$, indicating that JobRec learns a more globally consistent ranking that promotes jobs that are both appealing and feasible. 

\begin{table*}[t]
\centering
\small
\setlength{\tabcolsep}{6pt}
\begin{tabular}{l c c c c c c c}
\toprule
\textbf{Model} & \textbf{Eval} & \textbf{Recall@1} & \textbf{Recall@3} & \textbf{Recall@5} & \textbf{NDCG@1} & \textbf{NDCG@3} & \textbf{NDCG@5} \\
\midrule
\multirow{2}{*}{Preference-Only}
& Preference & 0.630 & 0.773 & 0.827 & 0.630 & 0.712 & 0.734 \\
& Qualification & 0.067 & 0.167 & 0.233 & 0.067 & 0.125 & 0.153 \\
\midrule
\multirow{2}{*}{\textbf{JobRec}}
& Preference & \textbf{0.468} & \textbf{0.672} & \textbf{0.752} & \textbf{0.468} & \textbf{0.587} & \textbf{0.620} \\
& Qualification & \textbf{0.333} & \textbf{0.533} & \textbf{0.767} & \textbf{0.333} & \textbf{0.451} & \textbf{0.548} \\
\bottomrule
\end{tabular}
\vspace{-1.5mm}
\caption{A diagnostic comparison between preference-only fine-tuning and JobRec, evaluated on both Preference and Qualification. Preference-only training achieves strong preference ranking but fails on qualification-aware retrieval, motivating explicit de-conflation.}
\label{tab:deconflation_necessary_full}
\vspace{-3mm}
\end{table*}

\subsection{RQ3: Why De-conflation is Necessary?}
Table~\ref{tab:deconflation_necessary_full} isolates the effect of conflating the two perspectives by fine-tuning the backbone with preference supervision only and evaluating the same preference head on both test sets. This model achieves strong preference ranking, with R@5 $=0.827$, indicating that single-sided supervision quickly fits what candidates like. However, it fails on qualification, where R@5 drops to $0.233$ under the same retrieval protocol. This gap reflects a structural mismatch: preference training emphasizes semantic desirability signals such as interests and role intent, while qualification requires evidence-based eligibility checks that are not learned from preference-only labels. JobRec preserves competitive preference performance while substantially improving qualification ranking, reaching Qual R@5 $=0.767$. This gain supports our formulation in Sec.~\ref{sec:decomposition}, where the observed success signal is a latent conjunction and negative outcomes are counterfactually ambiguous. Preference-only training effectively maps most failures to disinterest, entangling aspiration with eligibility and amplifying selection bias. 

\begin{table*}[t]
\centering
\small
\setlength{\tabcolsep}{4.5pt}
\begin{tabular}{l c c c c c c c}
\toprule
\textbf{Model / Stage} & \textbf{Eval} & \textbf{Recall@1} & \textbf{Recall@3} & \textbf{Recall@5} & \textbf{NDCG@1} & \textbf{NDCG@3} & \textbf{NDCG@5} \\
\midrule
\multirow{2}{*}{\textbf{Stage-1: Disentangled Pre-training}}
& Pref & 0.316 & 0.500 & 0.608 & 0.316 & 0.425 & 0.469 \\
& Qual & 0.233 & 0.433 & 0.500 & 0.233 & 0.342 & 0.368 \\
\midrule
\multirow{2}{*}{\textbf{Stage-1 + Stage-2: JobRec}}
& Pref & \textbf{0.468} & \textbf{0.672} & \textbf{0.752} & \textbf{0.468} & \textbf{0.587} & \textbf{0.620} \\
& Qual & \textbf{0.333} & \textbf{0.533} & \textbf{0.767} & \textbf{0.333} & \textbf{0.451} & \textbf{0.548} \\
\bottomrule
\end{tabular}
\vspace{-1.5mm}
\caption{\textbf{Effect of Stage-2 policy alignment}. Stage-2 consistently improves both preference and qualification ranking beyond Stage-1 disentangled pre-training.}
\label{tab:stage2_effect}
\vspace{-3mm}
\end{table*}

\subsection{RQ4: Effect of Stage-2 Policy Alignment}
Table~\ref{tab:stage2_effect} quantifies the contribution of Stage-2 policy alignment beyond Stage-1 disentangled pre-training. Stage-1 provides a solid foundation: separate heads for preference and qualification achieve moderate performance on both tasks, with Pref R@5 $=0.608$ and Qual R@5 $=0.500$. This indicates that de-conflation prevents the representation from collapsing into a single, ambiguous success signal. However, Stage-1 is purely predictive, estimating $s_{\text{pref}}$ and $s_{\text{qual}}$ independently without directly shaping the ranking policy that must balance interest and eligibility. Stage-2 closes this gap by optimizing the policy-level objective induced by the Lagrangian formulation in Sec.~\ref{sec:lagrangian}. After alignment, JobRec improves preference ranking from R@5 $0.608$ to $0.752$ and N@5 $0.469$ to $0.620$, and substantially strengthens qualification retrieval from R@5 $0.500$ to $0.767$ and N@5 $0.368$ to $0.548$. The largest gains appear at higher $K$, suggesting that Stage-2 improves the global ordering to better respect eligibility constraints rather than only correcting a few top positions. This matches the role of the dual variable, which calibrates the boundary so that highly preferred jobs are promoted only when supported by sufficient eligibility evidence.

\begin{figure}[t]
    \centering
    \includegraphics[width=0.45\textwidth]{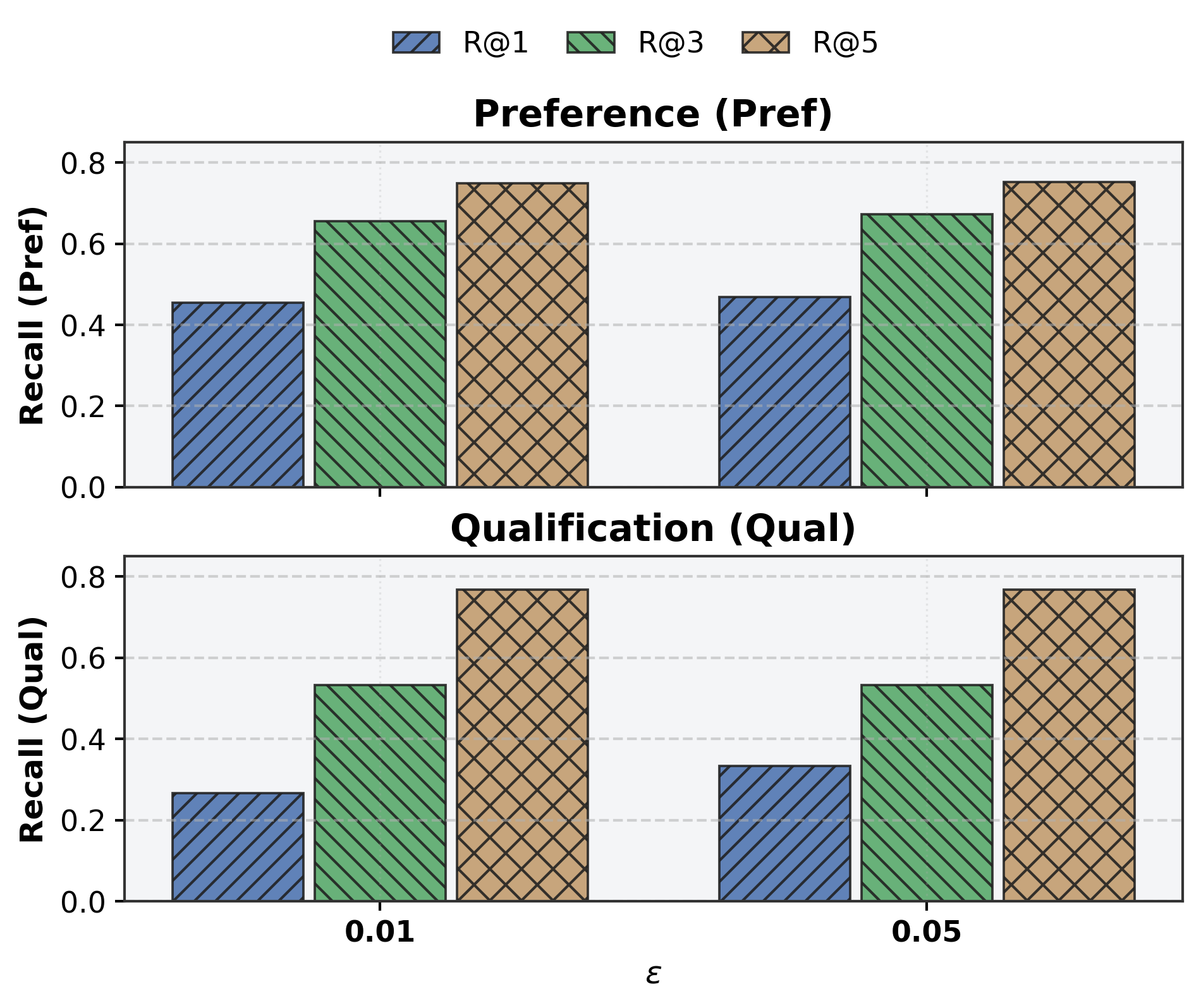}
    \caption{\textbf{Controllability analysis} by varying the eligibility target $\epsilon$ in our constraint-aligned policy. Larger $\epsilon$ enforces a stricter qualification requirement and yields improved qualification-aware ranking while maintaining stable preference performance.}
    \label{fig:control_eta}
\end{figure}

\subsection{RQ5: Controllability Analysis}
Figure~\ref{fig:control_eta} evaluates controllability by varying the eligibility target $\epsilon$, which sets the minimum qualification level enforced by policy alignment as defined in Sec.~\ref{sec:lagrangian}. Increasing $\epsilon$ from $0.01$ to $0.05$ yields a clear monotonic improvement in qualification ranking, most notably at the top of the list. Qual R@1 increases from $0.267$ to $0.333$, and Qual N@5 increases from $0.518$ to $0.548$, indicating that stricter eligibility targets promote more eligible jobs to higher ranks. Preference ranking remains stable and slightly improves, with Pref N@5 increasing from $0.610$ to $0.620$. This behavior matches the Lagrangian interpretation, where changing $\epsilon$ shifts the feasible boundary and moves the policy along the preference--eligibility trade-off rather than collapsing one objective. The fact that Qual R@5 and N@5 at $K=5$ remain unchanged while R@1 improves suggests that $\epsilon$ mainly controls the placement of eligible jobs near the top. 

\subsection{RQ6: How does JobRec reshape the preference ranking?}
\label{sec:rq6}

\begin{figure}[t]
    \centering
    \includegraphics[width=0.45\textwidth]{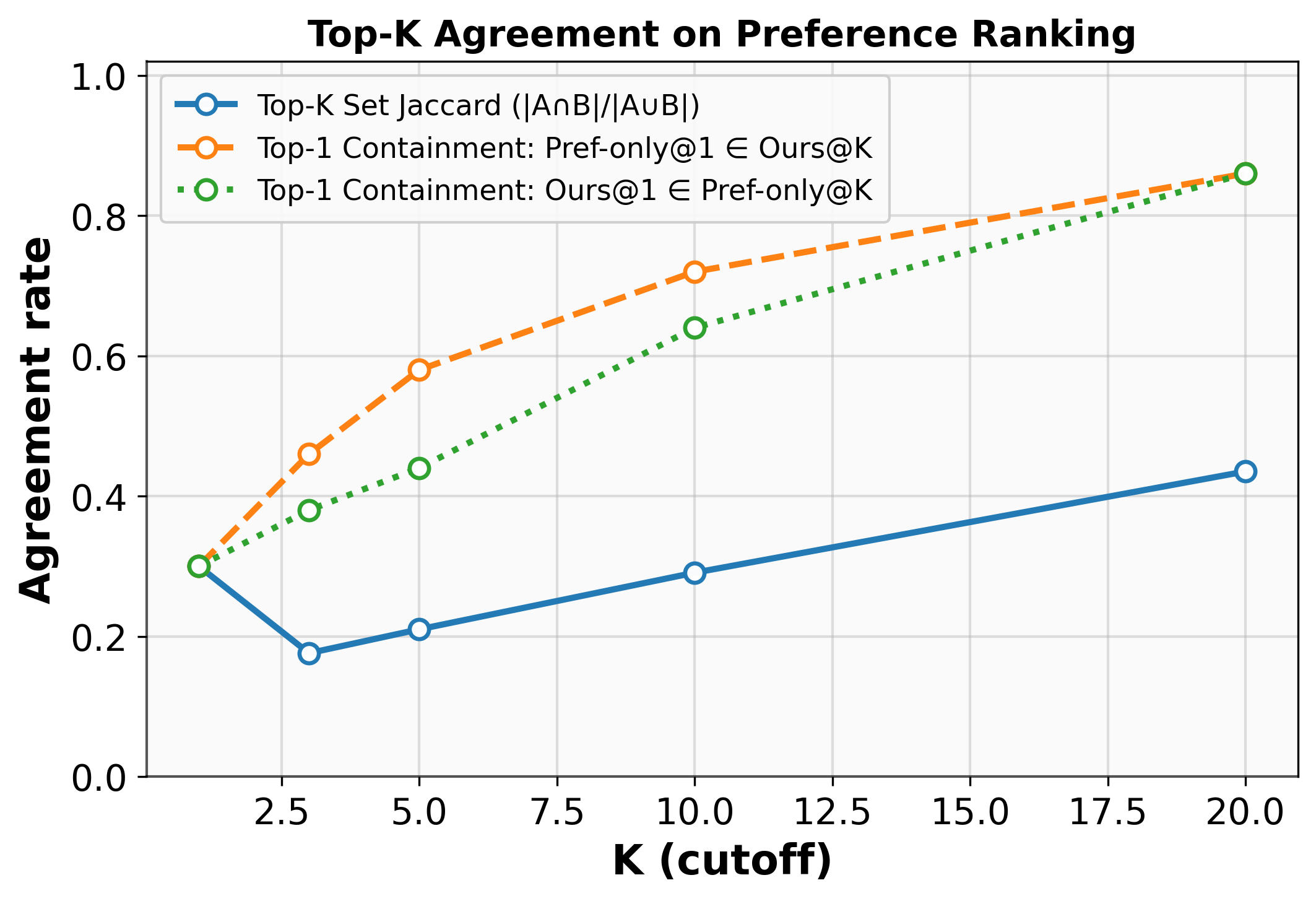}
    \caption{\textbf{Top-$K$ agreement on preference ranking.}
Using softmax-normalized $s_{\text{pref}}$, we measure per-user ranking consistency between \textsc{Pref-only} and \textsc{Ours} via Top-$K$ Jaccard overlap and bidirectional Top-1 containment (whether one method's Top-1 appears in the other's Top-$K$).}
\label{fig:topk_agreement_pref}
\end{figure}

Figure~\ref{fig:topk_agreement_pref} shows that our method produces a substantial yet structured change in preference ranking relative to Pref-only. After per-user softmax normalization, Top-$K$ Jaccard overlap remains low and increases slowly with $K$, indicating that our method changes the Top-$K$ set rather than simply rescaling scores. In contrast, Top-1 containment increases rapidly with $K$ and reaches about $0.86$ at $K=20$ in both directions. This pattern suggests that most disagreement concentrates at the top of the list: \textsc{JobRec} often replaces the baseline Top-1, but the replaced item typically stays within the other method's near-top set. The small-$K$ asymmetry further indicates conservative reranking, where \textsc{JobRec} preserves the baseline choice within its near-top candidates while promoting an alternative item to rank 1 when supported by our objective. 

\section{Conclusion}
In this paper, we propose \textbf{JobRec}, a dual-perspective LLM-based job recommendation framework that separates candidate preference from employer qualification and learns an eligibility-aware ranking policy via Lagrangian alignment. Built on USAS and a two-stage training strategy, JobRec consistently outperforms LLM-based and classical deep recommendation baselines on both objectives. Our analyses further establish the necessity of de-conflation, the benefits of Stage-two policy alignment, and practical controllability through the eligibility target. These results underscore the importance of modeling the two-sided decision structure for reliable matching.

\section*{Limitations}

Our study has several limitations that also indicate clear room for improvement. First, while USAS provides a structured interface for symmetric candidate--job representation, it is manually specified and its effectiveness depends on the quality of attribute extraction, which can be noisy in practice. Second, Stage-two alignment uses a global eligibility target with a shared dual variable, which yields an interpretable control knob but may be coarse when eligibility rules vary across employers, roles, or user segments. Third, the evaluation centers on a CS-domain benchmark with expert-refined synthetic supervision, which supports clean dual-perspective annotation but may underrepresent distribution shifts and policy variability present in real recruitment platforms. Overall, these constraints do not affect the core formulation. They motivate our next steps on broader domain coverage and robust schema and attribute extraction for deployment at scale.

\section*{Ethics Statement}
This paper studies LLM-based job recommendation under a dual-perspective formulation that separates candidate preference from employer qualification and enables eligibility-aware ranking. Our experiments use a CS-domain dataset constructed for research and expert refinement and do not include personally identifiable information. As with other LLM-based systems, outputs may reflect biases from pretraining data or schema and prompt design, and may be inaccurate. Practitioners should conduct bias auditing and apply human oversight in any real hiring context, since recommendations can affect equitable access to opportunities. We view JobRec as decision support rather than a stand-alone hiring decision maker, and responsible use should follow applicable policies and regulations.

\paragraph*{Use of AI assistants.}
We used ChatGPT as an AI writing assistant to polish grammar and improve clarity in the manuscript.
All technical content, experimental design, and reported results were produced and verified by the authors.


\bibliography{custom}

\begin{thebibliography}{61}
\providecommand{\natexlab}[1]{#1}

\bibitem[{Alsaif et~al.(2022)Alsaif, Sassi~Hidri, Eleraky, Ferjani, and Amami}]{alsaif2022learning}
Suleiman~Ali Alsaif, Minyar Sassi~Hidri, Hassan~Ahmed Eleraky, Imen Ferjani, and Rimah Amami. 2022.
\newblock Learning-based matched representation system for job recommendation.
\newblock \emph{Computers}, 11(11):161.

\bibitem[{Bao et~al.(2025)Bao, Zhang, Wang, Zhang, Yang, Luo, Chen, Feng, and Tian}]{bao2025bi}
Keqin Bao, Jizhi Zhang, Wenjie Wang, Yang Zhang, Zhengyi Yang, Yanchen Luo, Chong Chen, Fuli Feng, and Qi~Tian. 2025.
\newblock A bi-step grounding paradigm for large language models in recommendation systems.
\newblock \emph{ACM Transactions on Recommender Systems}, 3(4):1--27.

\bibitem[{Bao et~al.(2023)Bao, Zhang, Zhang, Wang, Feng, and He}]{bao2023tallrec}
Keqin Bao, Jizhi Zhang, Yang Zhang, Wenjie Wang, Fuli Feng, and Xiangnan He. 2023.
\newblock Tallrec: An effective and efficient tuning framework to align large language model with recommendation.
\newblock In \emph{Proceedings of the 17th ACM conference on recommender systems}, pages 1007--1014.

\bibitem[{{\c{C}}elik~Ertu{\u{g}}rul and Bitirim(2025)}]{ccelik2025job}
Duygu {\c{C}}elik~Ertu{\u{g}}rul and Selin Bitirim. 2025.
\newblock Job recommender systems: A systematic literature review, applications, open issues, and challenges.
\newblock \emph{Journal of Big Data}, 12(1):140.

\bibitem[{Chang et~al.(2025)Chang, Shi, Cao, Yang, Hwang, Wang, Pang, Wang, Liu, Peng et~al.}]{chang2025survey}
Ching Chang, Yidan Shi, Defu Cao, Wei Yang, Jeehyun Hwang, Haixin Wang, Jiacheng Pang, Wei Wang, Yan Liu, Wen-Chih Peng, and 1 others. 2025.
\newblock A survey of reasoning and agentic systems in time series with large language models.
\newblock \emph{arXiv preprint arXiv:2509.11575}.

\bibitem[{Chen et~al.(2026)Chen, Feng, Yang, Zhong, Shi, Li, Wei, Gao, Wu, Hu et~al.}]{chen2026self}
Yiqun Chen, Jinyuan Feng, Wei Yang, Meizhi Zhong, Zhengliang Shi, Rui Li, Xiaochi Wei, Yan Gao, Yi~Wu, Yao Hu, and 1 others. 2026.
\newblock Self-compression of chain-of-thought via multi-agent reinforcement learning.
\newblock \emph{arXiv preprint arXiv:2601.21919}.

\bibitem[{Chen et~al.(2025)Chen, Liu, Zhang, Sun, Ma, Yang, Shi, Mao, and Yin}]{chen2025tourrank}
Yiqun Chen, Qi~Liu, Yi~Zhang, Weiwei Sun, Xinyu Ma, Wei Yang, Daiting Shi, Jiaxin Mao, and Dawei Yin. 2025.
\newblock Tourrank: Utilizing large language models for documents ranking with a tournament-inspired strategy.
\newblock In \emph{Proceedings of the ACM on Web Conference 2025}, pages 1638--1652.

\bibitem[{Cui et~al.(2022)Cui, Ma, Zhou, Zhou, and Yang}]{cui2022m6}
Zeyu Cui, Jianxin Ma, Chang Zhou, Jingren Zhou, and Hongxia Yang. 2022.
\newblock M6-rec: Generative pretrained language models are open-ended recommender systems.
\newblock \emph{arXiv preprint arXiv:2205.08084}.

\bibitem[{Du et~al.(2024)Du, Luo, Yan, Wang, Liu, Zhu, Song, and Zhang}]{du2024enhancing}
Yingpeng Du, Di~Luo, Rui Yan, Xiaopei Wang, Hongzhi Liu, Hengshu Zhu, Yang Song, and Jie Zhang. 2024.
\newblock Enhancing job recommendation through {LLM}-based generative adversarial networks.
\newblock In \emph{Proceedings of the AAAI Conference on Artificial Intelligence}, volume~38, pages 8363--8371.

\bibitem[{Gu et~al.(2025)Gu, Zhong, Xia, Yang, Lu, Jiang, and Gai}]{gu2025r}
Hao Gu, Rui Zhong, Yu~Xia, Wei Yang, Chi Lu, Peng Jiang, and Kun Gai. 2025.
\newblock R 4ec: A reasoning, reflection, and refinement framework for recommendation systems.
\newblock In \emph{Proceedings of the Nineteenth ACM Conference on Recommender Systems}, pages 411--421.

\bibitem[{Guo et~al.(2017)Guo, Tang, Ye, Li, and He}]{guo2017deepfm}
Huifeng Guo, Ruiming Tang, Yunming Ye, Zhenguo Li, and Xiuqiang He. 2017.
\newblock Deepfm: a factorization-machine based neural network for ctr prediction.
\newblock \emph{arXiv preprint arXiv:1703.04247}.

\bibitem[{He et~al.(2020)He, Deng, Wang, Li, Zhang, and Wang}]{he2020lightgcn}
Xiangnan He, Kuan Deng, Xiang Wang, Yan Li, Yongdong Zhang, and Meng Wang. 2020.
\newblock Lightgcn: Simplifying and powering graph convolution network for recommendation.
\newblock In \emph{Proceedings of the 43rd International ACM SIGIR conference on research and development in Information Retrieval}, pages 639--648.

\bibitem[{Hu et~al.(2025)Hu, Lyu, Bai, and Cui}]{hu2025fairwork}
Yuhan Hu, Ziyu Lyu, Lu~Bai, and Lixin Cui. 2025.
\newblock Fairwork: A generic framework for evaluating fairness in llm-based job recommender system.
\newblock In \emph{Proceedings of the 48th International ACM SIGIR Conference on Research and Development in Information Retrieval}, pages 3964--3968.

\bibitem[{Jiang et~al.(2024)Jiang, Xia, Wei, Luo, Lin, and Huang}]{jiang2024diffmm}
Yangqin Jiang, Lianghao Xia, Wei Wei, Da~Luo, Kangyi Lin, and Chao Huang. 2024.
\newblock Diffmm: Multi-modal diffusion model for recommendation.
\newblock In \emph{Proceedings of the 32nd ACM International Conference on Multimedia}, pages 7591--7599.

\bibitem[{Kim et~al.(2024)Kim, Kang, Choi, Kim, Yang, and Park}]{kim2024large}
Sein Kim, Hongseok Kang, Seungyoon Choi, Donghyun Kim, Minchul Yang, and Chanyoung Park. 2024.
\newblock Large language models meet collaborative filtering: An efficient all-round llm-based recommender system.
\newblock In \emph{Proceedings of the 30th ACM SIGKDD Conference on Knowledge Discovery and Data Mining}, pages 1395--1406.

\bibitem[{Kokkodis and Ipeirotis(2023)}]{kokkodis2023good}
Marios Kokkodis and Panagiotis~G Ipeirotis. 2023.
\newblock The good, the bad, and the unhirable: Recommending job applicants in online labor markets.
\newblock \emph{Management Science}, 69(11):6969--6987.

\bibitem[{Kokkodis and Ransbotham(2023)}]{kokkodis2023learning}
Marios Kokkodis and Sam Ransbotham. 2023.
\newblock Learning to successfully hire in online labor markets.
\newblock \emph{Management Science}, 69(3):1597--1614.

\bibitem[{Kurek et~al.(2024)Kurek, Latkowski, Bukowski, {\'S}widerski, {\L}{\k{e}}picki, Baranik, Nowak, Zakowicz, and Dobrakowski}]{kurek2024zero}
Jaros{\l}aw Kurek, Tomasz Latkowski, Micha{\l} Bukowski, Bartosz {\'S}widerski, Mateusz {\L}{\k{e}}picki, Grzegorz Baranik, Bogusz Nowak, Robert Zakowicz, and {\L}ukasz Dobrakowski. 2024.
\newblock Zero-shot recommendation {AI} models for efficient job--candidate matching in recruitment process.
\newblock \emph{Applied Sciences}, 14(6):2601.

\bibitem[{Li et~al.(2023)Li, Wang, Li, Fu, Shen, Shang, and McAuley}]{li2023text}
Jiacheng Li, Ming Wang, Jin Li, Jinmiao Fu, Xin Shen, Jingbo Shang, and Julian McAuley. 2023.
\newblock Text is all you need: Learning language representations for sequential recommendation.
\newblock In \emph{Proceedings of the 29th ACM SIGKDD Conference on Knowledge Discovery and Data Mining}, pages 1258--1267.

\bibitem[{Li et~al.(2025)Li, Yang, Zhang, Xiao, Cao, Qin, Zhang, Zhao, and Bogdan}]{li2025climatellm}
Shixuan Li, Wei Yang, Peiyu Zhang, Xiongye Xiao, Defu Cao, Yuehan Qin, Xiaole Zhang, Yue Zhao, and Paul Bogdan. 2025.
\newblock Climatellm: Efficient weather forecasting via frequency-aware large language models.
\newblock \emph{arXiv preprint arXiv:2502.11059}.

\bibitem[{Lian et~al.(2018)Lian, Zhou, Zhang, Chen, Xie, and Sun}]{lian2018xdeepfm}
Jianxun Lian, Xiaohuan Zhou, Fuzheng Zhang, Zhongxia Chen, Xing Xie, and Guangzhong Sun. 2018.
\newblock xdeepfm: Combining explicit and implicit feature interactions for recommender systems.
\newblock In \emph{Proceedings of the 24th ACM SIGKDD International Conference on Knowledge Discovery \& Data Mining}, pages 1754--1763.

\bibitem[{Liao et~al.(2024)Liao, Li, Yang, Wu, Yuan, Wang, and He}]{liao2024llara}
Jiayi Liao, Sihang Li, Zhengyi Yang, Jiancan Wu, Yancheng Yuan, Xiang Wang, and Xiangnan He. 2024.
\newblock Llara: Large language-recommendation assistant.
\newblock In \emph{Proceedings of the 47th International ACM SIGIR Conference on Research and Development in Information Retrieval}, pages 1785--1795.

\bibitem[{Mao et~al.(2021)Mao, Zhu, Wang, Dai, Dong, Xiao, and He}]{mao2021simplex}
Kelong Mao, Jieming Zhu, Jinpeng Wang, Quanyu Dai, Zhenhua Dong, Xi~Xiao, and Xiuqiang He. 2021.
\newblock Simplex: A simple and strong baseline for collaborative filtering.
\newblock In \emph{Proceedings of the 30th ACM international conference on information \& knowledge management}, pages 1243--1252.

\bibitem[{Mulay et~al.(2022)Mulay, Sutar, Patel, Chhabria, and Mumbaikar}]{mulay2022job}
Aneesh Mulay, Shriyash Sutar, Jiten Patel, Aditi Chhabria, and Snehal Mumbaikar. 2022.
\newblock Job recommendation system using hybrid filtering.
\newblock In \emph{ITM Web of Conferences}, volume~44, page 02002. EDP Sciences.

\bibitem[{Naveed et~al.(2025)Naveed, Khan, Qiu, Saqib, Anwar, Usman, Akhtar, Barnes, and Mian}]{naveed2025comprehensive}
Humza Naveed, Asad~Ullah Khan, Shi Qiu, Muhammad Saqib, Saeed Anwar, Muhammad Usman, Naveed Akhtar, Nick Barnes, and Ajmal Mian. 2025.
\newblock A comprehensive overview of large language models.
\newblock \emph{ACM Transactions on Intelligent Systems and Technology}, 16(5):1--72.

\bibitem[{Ong and Khong(2025)}]{ong2025spectrum}
Rongqing~Kenneth Ong and Andy~WH Khong. 2025.
\newblock Spectrum-based modality representation fusion graph convolutional network for multimodal recommendation.
\newblock In \emph{Proceedings of the Eighteenth ACM International Conference on Web Search and Data Mining}, pages 773--781.

\bibitem[{Ping et~al.(2025)Ping, Li, Zhang, Cheng, Duan, Kanakaris, Xiao, Yang, Nazarian, Irimia et~al.}]{ping2025hdlcore}
Heng Ping, Shixuan Li, Peiyu Zhang, Anzhe Cheng, Shukai Duan, Nikos Kanakaris, Xiongye Xiao, Wei Yang, Shahin Nazarian, Andrei Irimia, and 1 others. 2025.
\newblock Hdlcore: A training-free framework for mitigating hallucinations in llm-generated hdl.
\newblock \emph{arXiv preprint arXiv:2503.16528}.

\bibitem[{Ren et~al.(2024{\natexlab{a}})Ren, Wei, Xia, Su, Cheng, Wang, Yin, and Huang}]{ren2024representation}
Xubin Ren, Wei Wei, Lianghao Xia, Lixin Su, Suqi Cheng, Junfeng Wang, Dawei Yin, and Chao Huang. 2024{\natexlab{a}}.
\newblock Representation learning with large language models for recommendation.
\newblock In \emph{Proceedings of the ACM web conference 2024}, pages 3464--3475.

\bibitem[{Ren et~al.(2024{\natexlab{b}})Ren, Chen, Yang, Li, Jiang, Cheng, Zhang, Mo, and Zhou}]{ren2024enhancing}
Yankun Ren, Zhongde Chen, Xinxing Yang, Longfei Li, Cong Jiang, Lei Cheng, Bo~Zhang, Linjian Mo, and Jun Zhou. 2024{\natexlab{b}}.
\newblock Enhancing sequential recommenders with augmented knowledge from aligned large language models.
\newblock In \emph{Proceedings of the 47th International ACM SIGIR Conference on Research and Development in Information Retrieval}, pages 345--354.

\bibitem[{Salinas et~al.(2023)Salinas, Shah, Huang, McCormack, and Morstatter}]{salinas2023unequal}
Abel Salinas, Parth Shah, Yuzhong Huang, Robert McCormack, and Fred Morstatter. 2023.
\newblock The unequal opportunities of large language models: Examining demographic biases in job recommendations by {C}hat{GPT} and {LLaMA}.
\newblock In \emph{Proceedings of the 3rd ACM Conference on Equity and Access in Algorithms, Mechanisms, and Optimization}, pages 1--15.

\bibitem[{Sun et~al.(2025)Sun, Ji, Zhu, Zhuang, He, and Xiong}]{sun2025market}
Ying Sun, Yang Ji, Hengshu Zhu, Fuzhen Zhuang, Qing He, and Hui Xiong. 2025.
\newblock Market-aware long-term job skill recommendation with explainable deep reinforcement learning.
\newblock \emph{ACM Transactions on Information Systems}, 43(2):1--35.

\bibitem[{Tan et~al.(2024)Tan, Xu, Hua, Ge, Li, and Zhang}]{tan2024idgenrec}
Juntao Tan, Shuyuan Xu, Wenyue Hua, Yingqiang Ge, Zelong Li, and Yongfeng Zhang. 2024.
\newblock Idgenrec: Llm-recsys alignment with textual id learning.
\newblock In \emph{Proceedings of the 47th international ACM SIGIR conference on research and development in information retrieval}, pages 355--364.

\bibitem[{Tavakoli et~al.(2022)Tavakoli, Faraji, Vrolijk, Molavi, Mol, and Kismih{\'o}k}]{tavakoli2022ai}
Mohammadreza Tavakoli, Abdolali Faraji, Jarno Vrolijk, Mohammadreza Molavi, Stefan~T Mol, and G{\'a}bor Kismih{\'o}k. 2022.
\newblock An {AI}-based open recommender system for personalized labor market driven education.
\newblock \emph{Advanced Engineering Informatics}, 52:101508.

\bibitem[{Touvron et~al.(2023)Touvron, Lavril, Izacard, Martinet, Lachaux, Lacroix, Rozi{\`e}re, Goyal, Hambro, Azhar et~al.}]{touvron2023llama}
Hugo Touvron, Thibaut Lavril, Gautier Izacard, Xavier Martinet, Marie-Anne Lachaux, Timoth{\'e}e Lacroix, Baptiste Rozi{\`e}re, Naman Goyal, Eric Hambro, Faisal Azhar, and 1 others. 2023.
\newblock Llama: Open and efficient foundation language models.
\newblock \emph{arXiv preprint arXiv:2302.13971}.

\bibitem[{Wang et~al.(2024{\natexlab{a}})Wang, Liu, Fan, Zhao, Kini, Yadav, Wang, Wen, Tang, and Liu}]{wang2024rethinking}
Hanbing Wang, Xiaorui Liu, Wenqi Fan, Xiangyu Zhao, Venkataramana Kini, Devendra Yadav, Fei Wang, Zhen Wen, Jiliang Tang, and Hui Liu. 2024{\natexlab{a}}.
\newblock Rethinking large language model architectures for sequential recommendations.
\newblock \emph{arXiv preprint arXiv:2402.09543}.

\bibitem[{Wang et~al.(2021)Wang, Shivanna, Cheng, Jain, Lin, Hong, and Chi}]{wang2021dcn}
Ruoxi Wang, Rakesh Shivanna, Derek Cheng, Sagar Jain, Dong Lin, Lichan Hong, and Ed~Chi. 2021.
\newblock Dcn v2: Improved deep \& cross network and practical lessons for web-scale learning to rank systems.
\newblock In \emph{Proceedings of the web conference 2021}, pages 1785--1797.

\bibitem[{Wang et~al.(2023)Wang, Jiang, Chen, Yang, Zhou, Cho, Fan, Huang, Lu, and Yang}]{wang2023recmind}
Yancheng Wang, Ziyan Jiang, Zheng Chen, Fan Yang, Yingxue Zhou, Eunah Cho, Xing Fan, Xiaojiang Huang, Yanbin Lu, and Yingzhen Yang. 2023.
\newblock Recmind: Large language model powered agent for recommendation.
\newblock \emph{arXiv preprint arXiv:2308.14296}.

\bibitem[{Wang et~al.(2024{\natexlab{b}})Wang, Pan, Jia, Wang, Wang, Feng, Li, Jiang, and Zhao}]{wang2024pre}
Yuhao Wang, Junwei Pan, Pengyue Jia, Wanyu Wang, Maolin Wang, Zhixiang Feng, Xiaotian Li, Jie Jiang, and Xiangyu Zhao. 2024{\natexlab{b}}.
\newblock Pre-train, align, and disentangle: Empowering sequential recommendation with large language models.
\newblock \emph{arXiv preprint arXiv:2412.04107}.

\bibitem[{Wang et~al.(2024{\natexlab{c}})Wang, Shi, and Zhao}]{wang2024mllm4rec}
Yuxiang Wang, Xin Shi, and Xueqing Zhao. 2024{\natexlab{c}}.
\newblock Mllm4rec: multimodal information enhancing llm for sequential recommendation.
\newblock \emph{Journal of Intelligent Information Systems}, pages 1--17.

\bibitem[{Wei et~al.(2024)Wei, Ren, Tang, Wang, Su, Cheng, Wang, Yin, and Huang}]{wei2024llmrec}
Wei Wei, Xubin Ren, Jiabin Tang, Qinyong Wang, Lixin Su, Suqi Cheng, Junfeng Wang, Dawei Yin, and Chao Huang. 2024.
\newblock Llmrec: Large language models with graph augmentation for recommendation.
\newblock In \emph{Proceedings of the 17th ACM International Conference on Web Search and Data Mining}, pages 806--815.

\bibitem[{Wu et~al.(2024)Wu, Qiu, Zheng, Zhu, and Chen}]{wu2024exploring}
Likang Wu, Zhaopeng Qiu, Zhi Zheng, Hengshu Zhu, and Enhong Chen. 2024.
\newblock Exploring large language model for graph data understanding in online job recommendations.
\newblock In \emph{Proceedings of the AAAI Conference on Artificial Intelligence}, volume~38, pages 9178--9186.

\bibitem[{Xia et~al.(2025{\natexlab{a}})Xia, Zhong, Gu, Yang, Lu, Jiang, and Gai}]{xia2025hierarchical}
Yu~Xia, Rui Zhong, Hao Gu, Wei Yang, Chi Lu, Peng Jiang, and Kun Gai. 2025{\natexlab{a}}.
\newblock Hierarchical tree search-based user lifelong behavior modeling on large language model.
\newblock In \emph{Proceedings of the 48th International ACM SIGIR Conference on Research and Development in Information Retrieval}, pages 1758--1767.

\bibitem[{Xia et~al.(2025{\natexlab{b}})Xia, Zhong, Song, Yang, Wan, Cai, Lu, and Jiang}]{xia2025trackrec}
Yu~Xia, Rui Zhong, Zeyu Song, Wei Yang, Junchen Wan, Qingpeng Cai, Chi Lu, and Peng Jiang. 2025{\natexlab{b}}.
\newblock Trackrec: Iterative alternating feedback with chain-of-thought via preference alignment for recommendation.
\newblock \emph{arXiv preprint arXiv:2508.15388}.

\bibitem[{Yang et~al.(2022)Yang, Hou, Song, Zhang, Wen, and Zhao}]{yang2022modeling}
Chen Yang, Yupeng Hou, Yang Song, Tao Zhang, Ji-Rong Wen, and Wayne~Xin Zhao. 2022.
\newblock Modeling two-way selection preference for person-job fit.
\newblock In \emph{Proceedings of the 16th ACM Conference on Recommender Systems}, pages 102--112.

\bibitem[{Yang et~al.(2023{\natexlab{a}})Yang, Fang, Zhang, Wu, and Lu}]{yang2023modal}
Wei Yang, Zhengru Fang, Tianle Zhang, Shiguang Wu, and Chi Lu. 2023{\natexlab{a}}.
\newblock Modal-aware bias constrained contrastive learning for multimodal recommendation.
\newblock In \emph{Proceedings of the 31st ACM International Conference on Multimedia}, pages 6369--6378.

\bibitem[{Yang et~al.(2024)Yang, Huo, and Liu}]{yang2024enhancing}
Wei Yang, Tengfei Huo, and Zhiqiang Liu. 2024.
\newblock Enhancing transformer-based semantic matching for few-shot learning through weakly contrastive pre-training.
\newblock In \emph{Proceedings of the 32nd ACM International Conference on Multimedia}, pages 10611--10620.

\bibitem[{Yang et~al.(2023{\natexlab{b}})Yang, Huo, Liu, and Lu}]{yang2023based}
Wei Yang, Tengfei Huo, Zhiqiang Liu, and Chi Lu. 2023{\natexlab{b}}.
\newblock based multi-intention contrastive learning for recommendation.
\newblock In \emph{Proceedings of the 46th International ACM SIGIR Conference on Research and Development in Information Retrieval}, pages 2339--2343.

\bibitem[{Yang et~al.(2025{\natexlab{a}})Yang, Pang, Li, Bogdan, Tu, and Thomason}]{yang2025maestro}
Wei Yang, Jiacheng Pang, Shixuan Li, Paul Bogdan, Stephen Tu, and Jesse Thomason. 2025{\natexlab{a}}.
\newblock Maestro: Learning to collaborate via conditional listwise policy optimization for multi-agent llms.
\newblock \emph{arXiv preprint arXiv:2511.06134}.

\bibitem[{Yang and Thomason(2025)}]{yang2025learning}
Wei Yang and Jesse Thomason. 2025.
\newblock Learning to deliberate: Meta-policy collaboration for agentic llms with multi-agent reinforcement learning.
\newblock \emph{arXiv preprint arXiv:2509.03817}.

\bibitem[{Yang et~al.(2025{\natexlab{b}})Yang, Weng, Pang, Cao, Ping, Zhang, Li, Zhao, Yang, Wang et~al.}]{yang2025toward}
Wei Yang, Muyan Weng, Jiacheng Pang, Defu Cao, Heng Ping, Peiyu Zhang, Shixuan Li, Yue Zhao, Qiang Yang, Mengdi Wang, and 1 others. 2025{\natexlab{b}}.
\newblock Toward evolutionary intelligence: Llm-based agentic systems with multi-agent reinforcement learning.
\newblock \emph{Available at SSRN 5819182}.

\bibitem[{Yang et~al.(2023{\natexlab{c}})Yang, Yang, and Liu}]{yang2023multimodal}
Wei Yang, Jie Yang, and Yuan Liu. 2023{\natexlab{c}}.
\newblock Multimodal optimal transport knowledge distillation for cross-domain recommendation.
\newblock In \emph{Proceedings of the 32nd ACM International Conference on Information and Knowledge Management}, pages 2959--2968.

\bibitem[{Yang and Yang(2024)}]{yang2024multimodal}
Wei Yang and Qingchen Yang. 2024.
\newblock Multimodal-aware multi-intention learning for recommendation.
\newblock In \emph{Proceedings of the 32nd ACM International Conference on Multimedia}, pages 5663--5672.

\bibitem[{Ye et~al.(2024)Ye, Yang, Cao, Zhang, Tang, Cai, and Liu}]{ye2024domain}
Wen Ye, Wei Yang, Defu Cao, Yizhou Zhang, Lumingyuan Tang, Jie Cai, and Yan Liu. 2024.
\newblock Domain-oriented time series inference agents for reasoning and automated analysis.
\newblock \emph{arXiv preprint arXiv:2410.04047}.

\bibitem[{Yue et~al.(2023)Yue, Rabhi, Moreira, Wang, and Oldridge}]{yue2023llamarec}
Zhenrui Yue, Sara Rabhi, Gabriel de Souza~Pereira Moreira, Dong Wang, and Even Oldridge. 2023.
\newblock Llamarec: Two-stage recommendation using large language models for ranking.
\newblock \emph{arXiv preprint arXiv:2311.02089}.

\bibitem[{Zhang et~al.(2024{\natexlab{a}})Zhang, Zhang, Wu, Wu, Xu, Zhao, Gao, Hu, and Chen}]{zhang2024notellm}
Chao Zhang, Haoxin Zhang, Shiwei Wu, Di~Wu, Tong Xu, Xiangyu Zhao, Yan Gao, Yao Hu, and Enhong Chen. 2024{\natexlab{a}}.
\newblock Notellm-2: Multimodal large representation models for recommendation.
\newblock \emph{arXiv preprint arXiv:2405.16789}.

\bibitem[{Zhang et~al.(2024{\natexlab{b}})Zhang, Zhang, Wu, Wu, Xu, Zhao, Gao, Hu, and Chen}]{zhang2024notellm2}
Chao Zhang, Haoxin Zhang, Shiwei Wu, Di~Wu, Tong Xu, Xiangyu Zhao, Yan Gao, Yao Hu, and Enhong Chen. 2024{\natexlab{b}}.
\newblock Notellm-2: Multimodal large representation models for recommendation.
\newblock \emph{arXiv preprint arXiv:2405.16789}.

\bibitem[{Zhang et~al.(2023)Zhang, Xie, Hou, Zhao, Lin, and Wen}]{zhang2023recommendation}
Junjie Zhang, Ruobing Xie, Yupeng Hou, Xin Zhao, Leyu Lin, and Ji-Rong Wen. 2023.
\newblock Recommendation as instruction following: A large language model empowered recommendation approach.
\newblock \emph{ACM Transactions on Information Systems}.

\bibitem[{Zhao et~al.(2025)Zhao, Zhong, Zheng, Yang, Lu, Jin, Jiang, and Gai}]{zhao2025hierarchical}
Rui Zhao, Rui Zhong, Haoran Zheng, Wei Yang, Chi Lu, Beihong Jin, Peng Jiang, and Kun Gai. 2025.
\newblock Hierarchical sequence id representation of large language models for large-scale recommendation systems.
\newblock In \emph{Companion Proceedings of the ACM on Web Conference 2025}, pages 641--650.

\bibitem[{Zhao et~al.(2023)Zhao, Zhou, Li, Tang, Wang, Hou, Min, Zhang, Zhang, Dong et~al.}]{zhao2023survey}
Wayne~Xin Zhao, Kun Zhou, Junyi Li, Tianyi Tang, Xiaolei Wang, Yupeng Hou, Yingqian Min, Beichen Zhang, Junjie Zhang, Zican Dong, and 1 others. 2023.
\newblock A survey of large language models.
\newblock \emph{arXiv preprint arXiv:2303.18223}, 1(2).

\bibitem[{Zhao et~al.(2024)Zhao, Wu, Wang, Tang, Wang, and De~Rijke}]{zhao2024let}
Yuyue Zhao, Jiancan Wu, Xiang Wang, Wei Tang, Dingxian Wang, and Maarten De~Rijke. 2024.
\newblock Let me do it for you: Towards llm empowered recommendation via tool learning.
\newblock In \emph{Proceedings of the 47th International ACM SIGIR Conference on Research and Development in Information Retrieval}, pages 1796--1806.

\bibitem[{Zheng et~al.(2023)Zheng, Qiu, Hu, Wu, Zhu, and Xiong}]{zheng2023generative}
Zhi Zheng, Zhaopeng Qiu, Xiao Hu, Likang Wu, Hengshu Zhu, and Hui Xiong. 2023.
\newblock Generative job recommendations with large language model.
\newblock \emph{arXiv preprint arXiv:2307.02157}.

\end{thebibliography}

\appendix

\section{Related Work}
\label{app:appendix_relatedwork}

\subsection{Generative Recommendation}
\label{app:related_generative_rec}

Due to its powerful reasoning and standardization capabilities, Large Language Models (LLMs) are widely applied in various downstream tasks~\cite{ping2025hdlcore,li2025climatellm,ye2024domain,yang2025learning,chang2025survey,yang2025maestro,naveed2025comprehensive}. Generative recommendation reframes recommendation as language modeling: user histories and items are serialized into text (or language-like symbols) and solved via pre-trained LLMs~\cite{zhao2023survey,yang2024enhancing,yang2025toward,gu2025r}, aiming to improve transferability, cold-start robustness, and task unification\cite{li2023text, wang2024rethinking, bao2023tallrec}. Although some multimodal recommendation methods have achieved certain results~\cite{jiang2024diffmm,yang2023multimodal,yang2023modal,yang2023based,ong2025spectrum,yang2024multimodal}, generative models have prominent advantages in understanding semantics. Early efforts demonstrate that sequential recommendation can be cast into sentence-level modeling by textualizing item attributes and learning generalizable language representations \cite{li2023text}, and that a single pretrained language model can be adapted to support open-ended recommendation domains and tasks through text-based formulations \cite{ yue2023llamarec, wang2023recmind, zhang2023recommendation}.

A key challenge is the \emph{misalignment} between LLM semantic priors and recommender objectives, especially collaborative signals and the discrete item space\cite{ren2024enhancing, liao2024llara, wei2024llmrec,xia2025hierarchical}. Recent work therefore emphasizes principled alignment and grounding. Tuning-based frameworks adapt LLMs with recommendation data to close the task gap \cite{bao2023tallrec}, while PAD further couples pretraining with recommendation-anchored alignment and disentangled experts to better reconcile textual and collaborative views \cite{wang2024pre}. For item grounding, BIGRec proposes a bi-step paradigm that first teaches LLMs to generate item-related tokens and then maps them to actual items for comprehensive ranking \cite{bao2025bi}, and IDGenRec learns concise textual IDs to represent items in a platform-agnostic yet semantically meaningful way \cite{tan2024idgenrec}.

Another practical concern is inference efficiency. Instead of relying on full autoregressive generation, Lite-LLM4Rec streamlines LLM architectures by replacing decoding with direct ranking heads and designing hierarchical processing for long contexts \cite{wang2024rethinking}. Similarly, LlamaRec adopts a two-stage retrieval--reranking design and uses a verbalizer-style scoring mechanism to rank candidate items without generating long outputs \cite{yue2023llamarec}.

Beyond end-to-end generation, LLMs are increasingly used as \emph{augmenters} and \emph{interfaces} that complement conventional recommenders. LLMRec improves recommendation by augmenting interaction graphs with LLM-derived signals alongside denoising mechanisms \cite{wei2024llmrec}, while RLMRec enhances model-agnostic representation learning via LLM profiling and cross-view alignment between semantic and collaborative spaces \cite{ren2024representation}. A-LLMRec explicitly injects collaborative knowledge from pretrained CF recommenders into LLM-based pipelines to achieve strong performance in both cold and warm regimes with efficient adaptation \cite{kim2024large}. In retrieval-centric settings, NoteLLM and NoteLLM-2 learn LLM-based (multi)modal representations for item-to-item recommendation, using prompt-driven compression and modality-aware integration to obtain effective embeddings \cite{zhang2024notellm,zhang2024notellm2}. Finally, ToolRec views recommendation as a controllable decision process where an LLM invokes retrieval/ranking tools, mitigating hallucinations and improving preference elicitation through tool learning \cite{zhao2024let}.

Overall, existing studies suggest that successful generative recommendation hinges on: (i) effective text/symbol representations for behaviors and items \cite{li2023text,cui2022m6}, (ii) grounding and alignment to the discrete item and collaborative spaces \cite{bao2023tallrec,wang2024pre,bao2025bi,tan2024idgenrec}, and (iii) system designs that prioritize efficient ranking and modular integration with traditional recommenders \cite{wang2024rethinking,yue2023llamarec,wei2024llmrec,ren2024representation,kim2024large,xia2025trackrec,zhao2025hierarchical,zhao2024let}.

\subsection{Job Recommendation}
\label{app:related_job_rec}
Job recommender systems (JRS) match job seekers with vacancies on large-scale recruitment platforms under heterogeneous and often incomplete signals, including resumes, interaction logs, and job postings. Compared with conventional recommendation, JRS must satisfy hard constraints such as eligibility and skill requirements, handle rapid job turnover, and operate under higher stakes for both candidates and employers. These factors intensify sparsity and cold start and make robustness, fairness, and evaluation protocols central. A recent systematic review summarizes JRS paradigms and evaluation criteria beyond ranking accuracy, and highlights open challenges such as noisy profiles, evolving labor demand, and demographic bias \cite{ccelik2025job}.

Early JRS methods largely follow content-based, collaborative, and hybrid pipelines. Content-based systems align explicit features between resumes and job descriptions, often skill-centric, which yields interpretable but limited matching \cite{alsaif2022learning}. Hybrid approaches combine content and collaborative signals to mitigate overspecialization and sparsity \cite{mulay2022job}. Despite their practical appeal, these methods remain constrained by incomplete profiles, reliance on surface feature overlap, and limited ability to model multi-faceted preferences or longer-term trajectories.

Beyond immediate matching, recent work connects recommendation with labor-market dynamics and upskilling. Education-oriented approaches mine skill requirements from vacancies and recommend learning pathways and educational resources, linking demand to actionable training \cite{tavakoli2022ai}. Market-aware formulations model skill learning utilities and optimize long-term outcomes with multi-objective reinforcement learning and exemplar-based explanations \cite{sun2025market}. These studies emphasize long-horizon decision making and explainability, but focus more on skill roadmaps than ranking under noisy, partially observed profiles.

Recent progress increasingly leverages large language models to improve semantic understanding and compensate for missing or low-quality profiles. LLM-based pipelines have been used to complete resumes for improved recommendation, with additional mechanisms to mitigate hallucination by extracting reliable properties and refining representations through distribution alignment \cite{du2024enhancing}. Other work integrates behavior graphs via meta-path prompting and path augmentation to preserve graph semantics while reducing prompt bias from serialization \cite{wu2024exploring}. In parallel, pretrained encoders support zero-shot job--candidate matching as an efficient alternative to task-specific training \cite{kurek2024zero}. As LLMs enter the decision pipeline, fairness risks become more salient: intersectional demographic biases have been documented in LLM-driven job recommendations, motivating systematic auditing and mitigation in high-stakes settings \cite{salinas2023unequal}. Together, these works suggest that LLMs can strengthen JRS semantic profiling and reasoning, while raising new challenges in reliability, controllability, and equity.

\begin{figure*}[t]
    \centering
    \includegraphics[width=\textwidth]{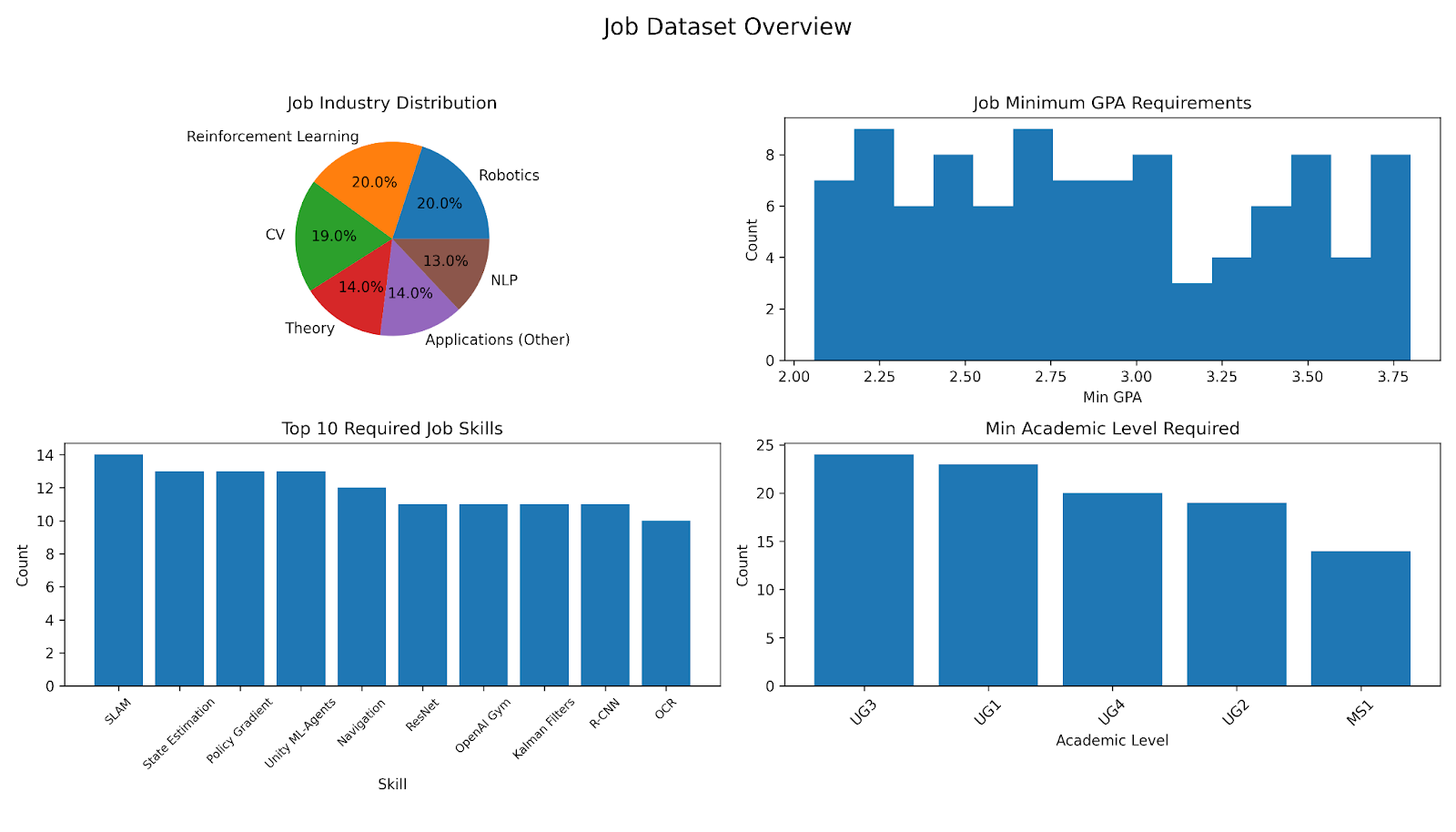}
    \caption{\textbf{Job dataset overview.} Summary statistics of the synthesized CS-domain job postings. The figure reports the distribution of job industries (top-left), the distribution of minimum GPA requirements (top-right), the top required skills across postings (bottom-left), and the minimum academic level requirements (bottom-right).}
\label{fig:data1}
\end{figure*}

\section{Experiment}
\label{app:appendix_exp}

\subsection{Experimental Setup}
\label{app:appendix_exp_setup}

\paragraph{Dataset and Evaluation Protocol.}
The dataset is generated by synthesizing user/job profiles and then applying an \textbf{internal scoring rubric} to all user--job combinations to obtain a compatibility score for each pair, which is subsequently used to derive task labels (details in Sec.~\ref{app:appendix_exp_syndata}). For evaluation, we adopt a sampled ranking protocol. For the preference task, for each ground-truth positive interaction in the test set, we form a candidate set by pairing it with 49 hard negative jobs per positive, and rank the resulting set according to each method's predicted score. For the qualification task, candidate sets are job-centric with variable batch sizes depending on the number of applicants per job. Hard negatives are sampled within the corresponding split to avoid information leakage across train/validation/test.

\paragraph{Data splits and training subset.}
We split the dataset into training/validation/test sets with a ratio of $7{:}1{:}2$ for the \textbf{preference} task (LABEL\_PREF).

Due to training time constraints, we train preference models using a randomly sampled subset of the preference-training split: \textbf{68 batches} (each batch contains 1 positive and 49 hard negatives, i.e., 50 candidates per query). The preference test set uses \textbf{1,402 batches}. This subsetting is used only for training efficiency; evaluation is performed on the full test split.

For the \textbf{qualification} task (LABEL\_QUAL), the dataset is job-centric and is constructed from the preference-positive pool (details in Sec.~\ref{app:appendix_exp_syndata}). Because there are only $100$ jobs, we split by jobs into \textbf{68 jobs for training} and \textbf{30 jobs for testing} (i.e., $7{:}3$), and do not allocate a separate validation split.

\paragraph{Baselines.} JobRec uses a Llama-3-8B~\cite{touvron2023llama} backbone fine-tuned with LoRA using learning rate $5\times 10^{-5}$, rank $r{=}16$, $\alpha{=}32$, and dropout $0.10$ applied to $\{Q,K,V,O\}$ modules, and we select checkpoints based on validation performance. We compare against LLM baselines including Zero-Shot Prompting, In-Context Learning with four demonstrations, and TallRec, as well as non-LLM recommenders including SimpleX~\cite{mao2021simplex}, LightGCN~\cite{he2020lightgcn}, DeepFM~\cite{guo2017deepfm}, xDeepFM~\cite{lian2018xdeepfm} and DCNv2~\cite{wang2021dcn}.

\begin{figure*}[t]
    \centering
    \includegraphics[width=\textwidth]{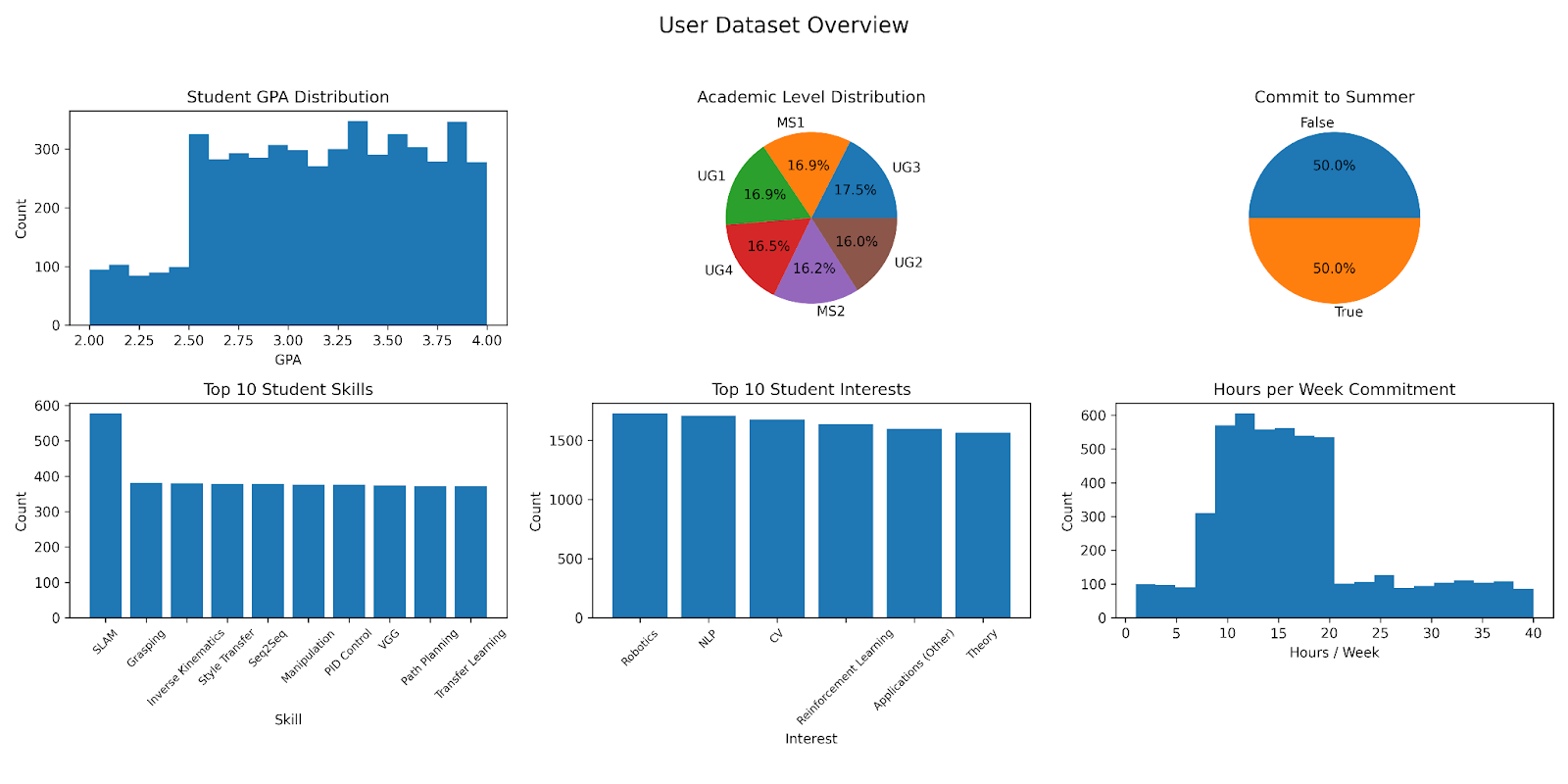}
    \caption{\textbf{User dataset overview.} Summary statistics of the synthesized CS-domain candidate profiles. The figure reports the GPA distribution (top-left), the distribution of academic levels (top-middle), the distribution of summer availability commitments (top-right), the top skills (bottom-left), the top research and career interests (bottom-middle), and the weekly time commitment distribution in hours (bottom-right).}
\label{fig:data2}
\end{figure*}

\subsection{Synthetic Data}
\label{app:appendix_exp_syndata}

\paragraph{Data source, privacy, and sensitive content.}
All candidate profiles and job descriptions in our benchmark are \textbf{fully synthetic} and generated under the USAS-guided synthesis protocol.
We \textbf{do not use, scrape, or derive} any content from real resumes, real job postings, or platform interaction logs.
Accordingly, the released data contains \textbf{no personally identifying information} and is not intended to represent any specific individual.
We also include basic safety checks during generation and review to avoid personal names, contact information, and offensive content.
Since the dataset is synthetic by construction, there is no linkage to real users, and no user privacy risks are introduced by data collection. Dataset statistics are summarized in Figures~\ref{fig:data1} and~\ref{fig:data2}.

\paragraph{Human involvement and roles.}
We involve domain experts in an expert-in-the-loop refinement stage.
The experts are computer science researchers, including ten PhD-level participants and faculty-level researchers, with backgrounds spanning machine learning, deep learning, and reinforcement learning.
Their role is to \textbf{audit and correct} a subset of automatically assigned silver labels, focusing on technical plausibility and boundary cases that require domain judgment.
Experts do not provide personal data, do not access any private user information, and operate solely on synthetic candidate--job pairs.

\paragraph{Instructions and consent.}
Experts are instructed to review only the information shown in each synthetic candidate--job pair and to assign or correct labels for preference and qualification based on the stated requirements and evidence.
They are instructed \textbf{not to introduce real-world personal details} and to flag any sample that appears to contain sensitive or identifying content.
All participants agree to take part in this annotation and review process under these constraints.

\paragraph{Recruitment and compensation.}
Experts participate on a voluntary, pro bono basis through academic collaboration channels.

\paragraph{Ethics review board approval.}
Because the dataset is fully synthetic and the expert review does not involve collecting personal data from human subjects, this process does not constitute human-subject data collection from individuals represented in the dataset.
The expert review is conducted as a standard research collaboration activity on synthetic materials.

\paragraph{Release plan.}
We plan to release the dataset and accompanying code to support reproducibility.
The release includes the synthetic candidate--job pairs, dual-perspective labels, and documentation of the generation and expert refinement protocol.

\paragraph{Profile generation.}
The full dataset contains $5{,}000$ users and $100$ jobs. User profiles are generated across six categories of Computer Science (Robotics, CV, NLP, Applications, Reinforcement Learning, and Theory). Users and jobs are created independently, with users described by 21 structured features and jobs by 12 structured features.

\paragraph{Rubric-based scoring.}
We enumerate all user--job combinations and score each pair using an internal rubric to produce a scalar compatibility score.

\paragraph{Preference labels (LABEL\_PREF).}
Preference labels are determined directly from the rubric score: scores $\ge 14$ are labeled as positive, scores in $[10,14)$ are labeled as hard negatives, and scores below 10 are discarded to focus evaluation on challenging near-miss cases. For splitting, we first split only the positives and apply $k$-core filtering (3:1:1) such that each user appears 3 times in train, 1 time in validation, and 1 time in test. We then sample 49 hard negatives per positive within each split (from the $[10,14)$ pool) for that user to form a 50-item candidate set.

\paragraph{Qualification labels (LABEL\_QUAL) derived from preference positives.}
The qualification dataset is generated \emph{via the positives of the preference dataset}. After scoring, we transform the data into a job-centric format: for each job, we collect all users with rubric score $\ge 14$ as \emph{applicants} (i.e., users who would ``prefer'' the job under the rubric). Among these applicants, the single applicant with the highest score is labeled as qualified ($Y_{\text{qual}}=1$), while all other applicants are labeled as not qualified ($Y_{\text{qual}}{=}0$). This yields varying batch sizes per job depending on the number of applicants. This top-1 labeling emulates shortlist-style screening where only a small fraction of applicants pass employer-side eligibility checks.

\paragraph{Final splits used in our experiments.}
After filtering, scoring, and subset selection, the final experimental splits are:
(i) LABEL\_PREF: \textbf{68 batches for training} (subset of the original preference-train split) and \textbf{1,402 batches for testing};
(ii) LABEL\_QUAL: \textbf{68 jobs/batches for training} and \textbf{30 jobs/batches for testing}.

\subsection{Experiment: How does JobRec reshape the preference ranking?}
\label{app:rq6}

To isolate the effect of our training objective on the preference score, we compare rankings induced by $s_{\mathrm{pref}}$ from \textsc{Pref-only} and \textsc{Ours}. For each user, we compute the Top-$K$ set under each method and report two measures. The first is the Top-$K$ set Jaccard coefficient, defined as $\frac{|A\cap B|}{|A\cup B|}$, which quantifies overlap between the two Top-$K$ sets. The second is Top-1 containment, which checks whether the Top-1 item from one method appears in the other method's Top-$K$ set. We apply per-user softmax normalization to the scores, so these measures reflect ranking changes rather than score-scale effects.

\paragraph{Preference rankings change meaningfully rather than being preserved.}
Figure~\ref{fig:topk_agreement_pref} shows that the Top-$K$ Jaccard overlap remains low across $K$, staying around $0.2$ to $0.45$ as $K$ increases. This indicates that our method does not simply rescale the same candidate set, but changes the ordering of the preference list. This behavior matches our objective, which reallocates probability mass within each user so that the induced ranking better supports the overall decision criterion.

\paragraph{Top-1 changes are common but remain locally consistent.}
Although the Top-$K$ sets differ, the containment curves show that the Top-1 item from either method increasingly appears in the other method's Top-$K$ set as $K$ grows, reaching about $0.86$ at $K=20$ in both directions. This suggests that disagreements are concentrated at the head of the list. \textsc{Ours} often replaces the top-ranked item from \textsc{Pref-only}, but the replaced item typically remains near the top under the alternative ranking rather than being far from it. This pattern indicates targeted head-of-list adjustments instead of broad reshuffling.

\paragraph{Small-$K$ asymmetry indicates conservative reranking at the head.}
At small cutoffs, containment is asymmetric. The Top-1 item from \textsc{Pref-only} more often appears in \textsc{Ours} Top-$K$ than the reverse for $K=3$ and $K=5$. This suggests that \textsc{Ours} tends to keep the baseline top choice among its near-top candidates, while selecting a different item as Top-1 when supported by the training signal. Overall, the method acts as a controlled reranker that changes the final decision when it improves the objective, while limiting disruptive drift in the preference ordering.

\end{document}